%% file: main.tex
\documentclass[12pt]{article}
\usepackage[english]{babel}
\usepackage[utf8]{inputenc}
\usepackage{johd}

\usepackage{microtype}
\usepackage{graphicx}
\usepackage{subfigure}
\usepackage{booktabs} 

\usepackage{hyperref}

\usepackage{tikz}
\usetikzlibrary{external}
\usepackage[utf8]{inputenc}
\usepackage{pgfplots}
\DeclareUnicodeCharacter{2212}{−}
\usepgfplotslibrary{groupplots,dateplot}
\usetikzlibrary{patterns,shapes.arrows}
\pgfplotsset{compat=newest}
\usepackage{overpic}

\usepackage{amsmath}
\usepackage{amssymb}
\usepackage{mathtools}
\usepackage{amsthm}

\usepackage[capitalize,noabbrev]{cleveref}

\theoremstyle{plain}

\theoremstyle{definition}

\theoremstyle{remark}

\newcommand{\norm}[1]{\left\lVert#1\right\rVert}
\DeclarePairedDelimiter\abs{\lvert}{\rvert}
\DeclareMathOperator*{\argmin}{\arg\!\min}

\usepackage{cuted}

\usepackage[textsize=tiny]{todonotes}
\usepackage[textsize=tiny]{todonotes}
\usepackage{svg}
\svgpath{{images/}}

\usepackage{subfigure}
\usepackage{mathtools}

\title{3D Shape Completion with Test-Time Training}

\author{Michael Schopf-Kuester$^{1}$, Zorah L\"{a}hner$^{1,2,3}$, Michael Moeller$^{1}$ \\
        \small $^{1}$University of Siegen, Germany,
        \small $^{2}$University of Bonn, Germany,\\
        \small $^{3}$Lamarr Institute, Germany \\\\
        \small Corresponding author: Michael Schopf-Kuester; 
        \tt{michael.schopf@uni-siegen.de}
}
\date{}
\begin{document}
\maketitle
\begin{abstract} 
\noindent This work addresses the problem of \textit{shape completion}, i.e., the task of restoring incomplete shapes by predicting their missing parts. While previous works have often predicted the fractured and restored shape in one step, we approach the task by separately predicting the fractured and newly restored parts, but ensuring these predictions are interconnected. We use a decoder network motivated by related work on the prediction of signed distance functions (DeepSDF). In particular, our representation allows us to consider \textit{test-time-training}, i.e., finetuning network parameters to match the given incomplete shape more accurately during inference. While previous works often have difficulties with artifacts around the fracture boundary, we demonstrate that our overfitting to the fractured parts leads to significant improvements in the restoration of eight different shape categories of the ShapeNet data set in terms of their chamfer distances. \end{abstract}

\begin{figure*}
  \centering
  \begin{tabular}[b]{c}
  \begin{overpic}[trim={600 100 600 100},clip,width=.14\linewidth,keepaspectratio]{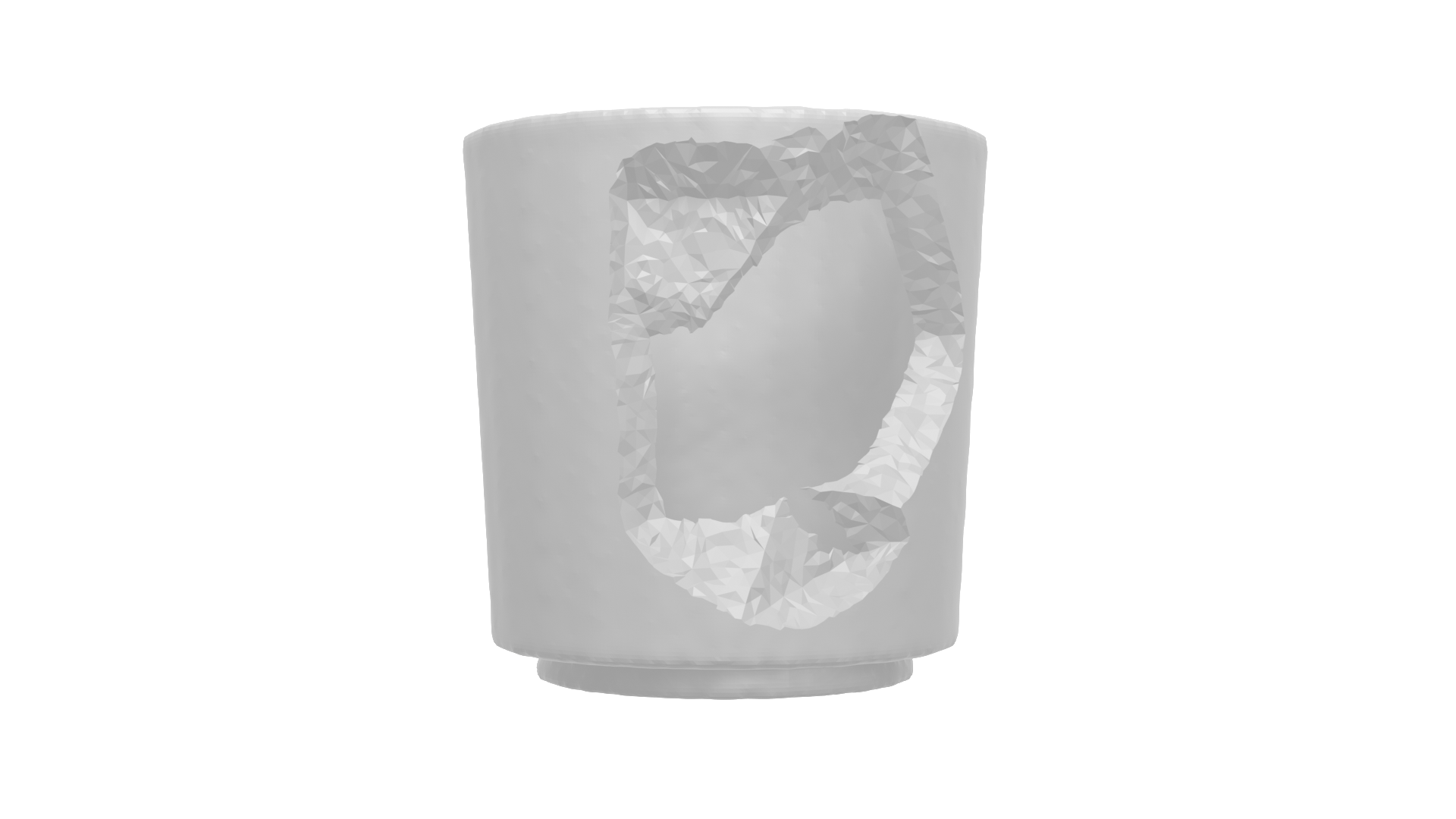}
  \put(20,-1){\tiny Fractured Input}
  \end{overpic}
  \begin{overpic}[trim={700 220 700 290},clip,width=.14\linewidth]{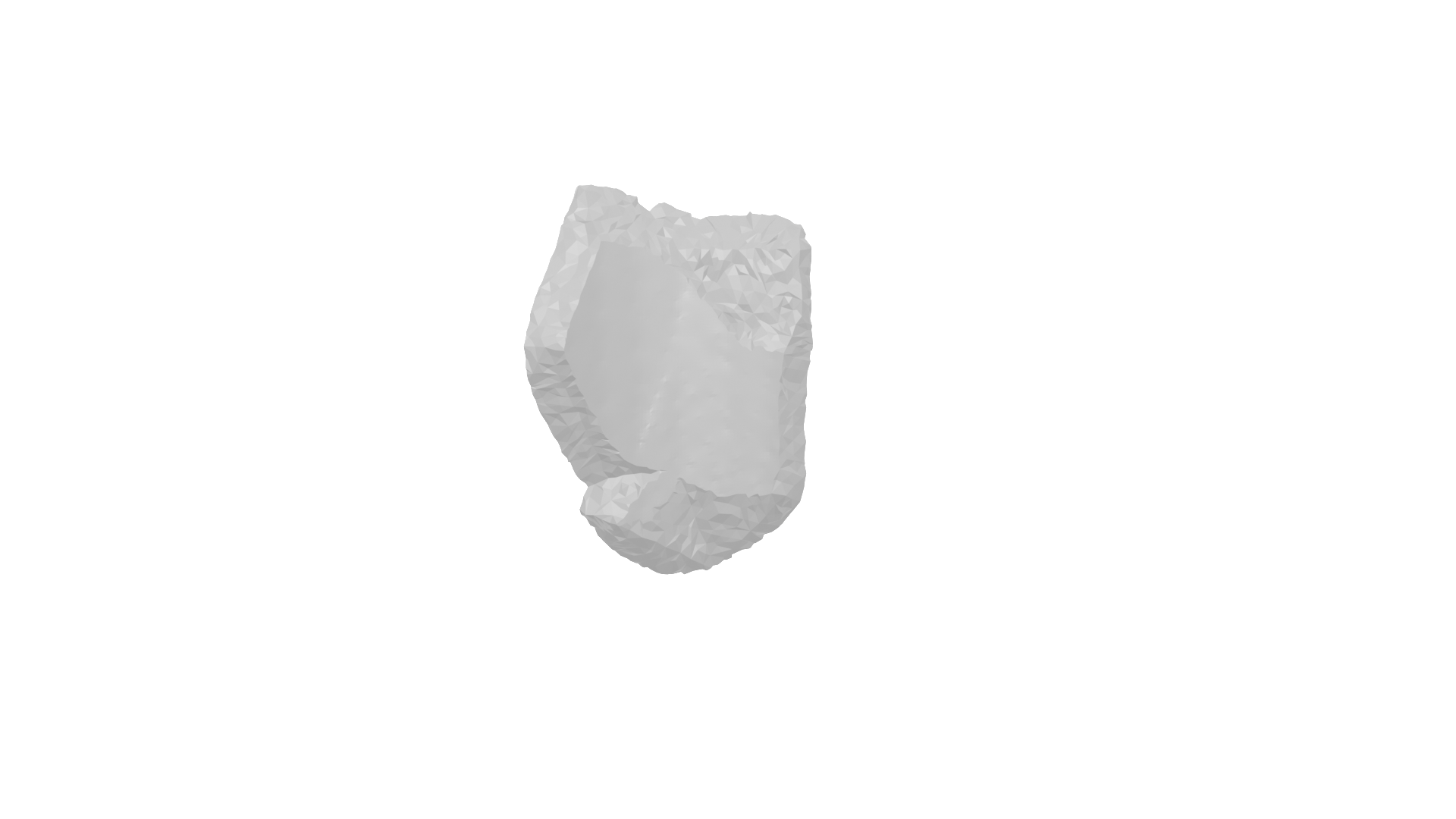}
  \put(0,-1){\tiny Missing Restoration Part}
  \end{overpic}\\
  \begin{overpic}[trim={600 100 600 100},clip,width=.14\linewidth]{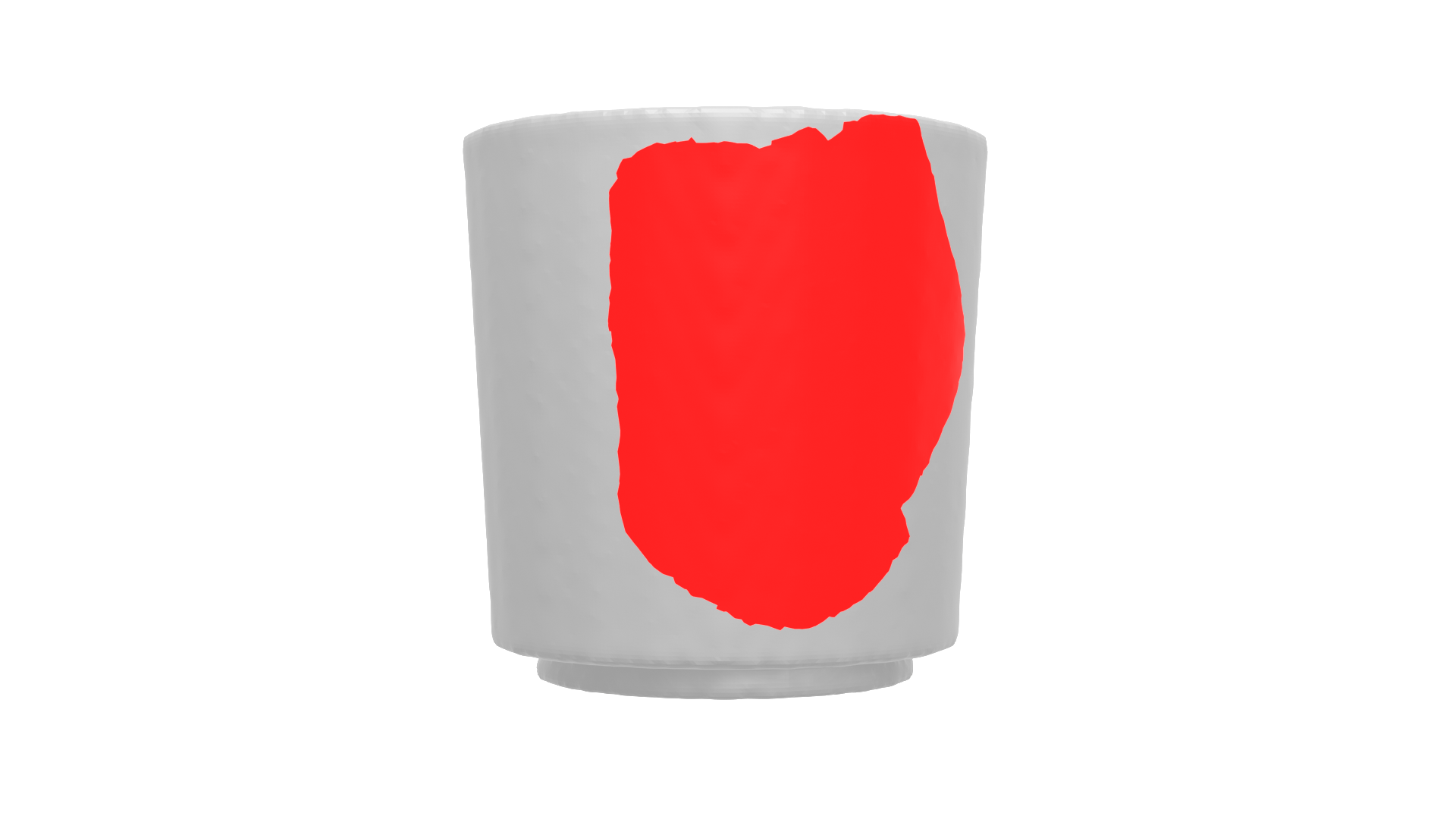} 
  \put(30,0){\tiny Combined}
    \end{overpic}\\
    \small (a) Ground Truth
  \end{tabular}~
  \begin{tabular}[b]{c}
    \includegraphics[trim={600 100 600 100},clip,width=.14\linewidth]{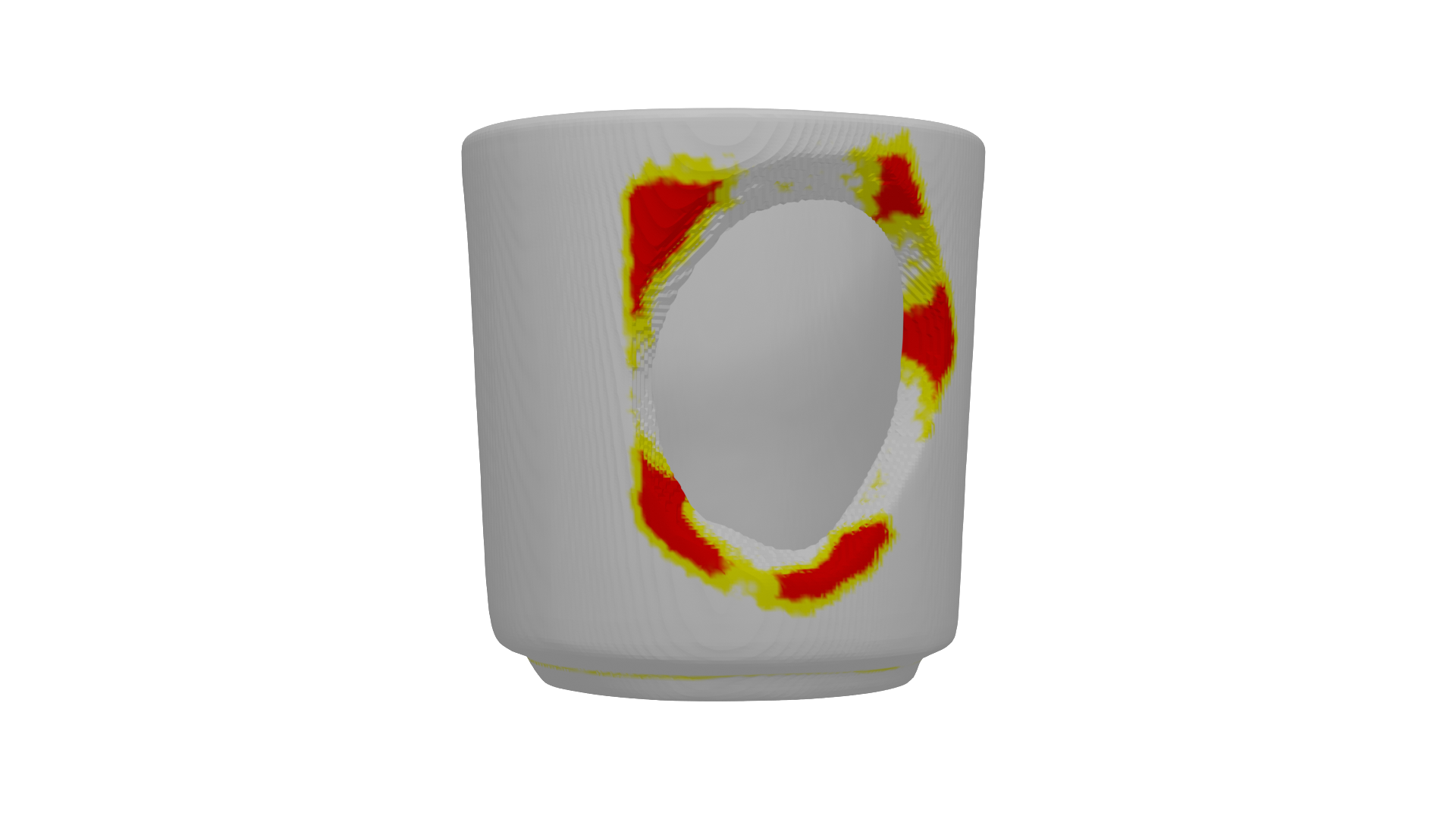} 
    \includegraphics[trim={700 220 700 290},clip,width=.14\linewidth]{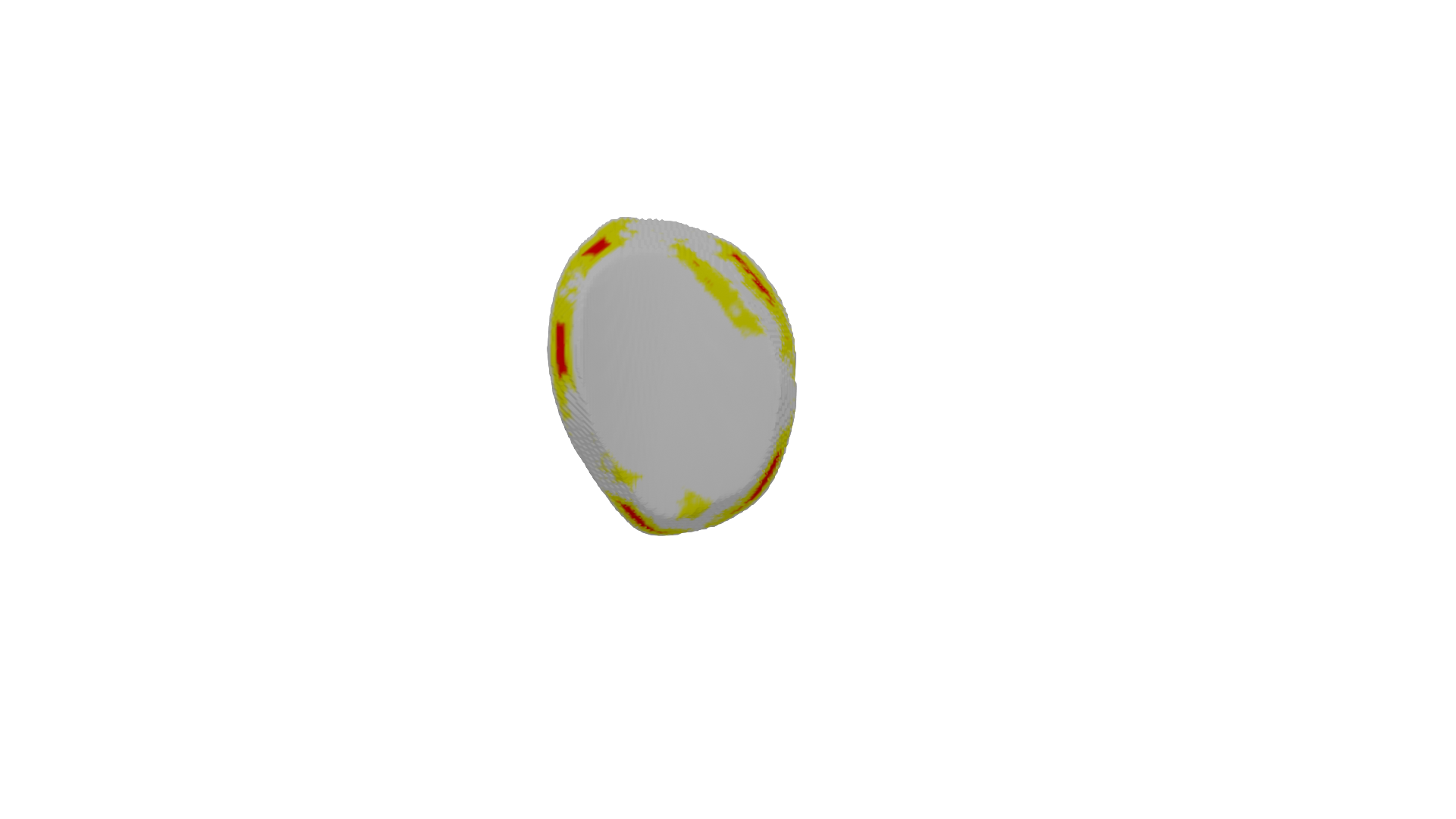}\\
    \includegraphics[trim={600 100 600 100},clip,width=.14\linewidth]{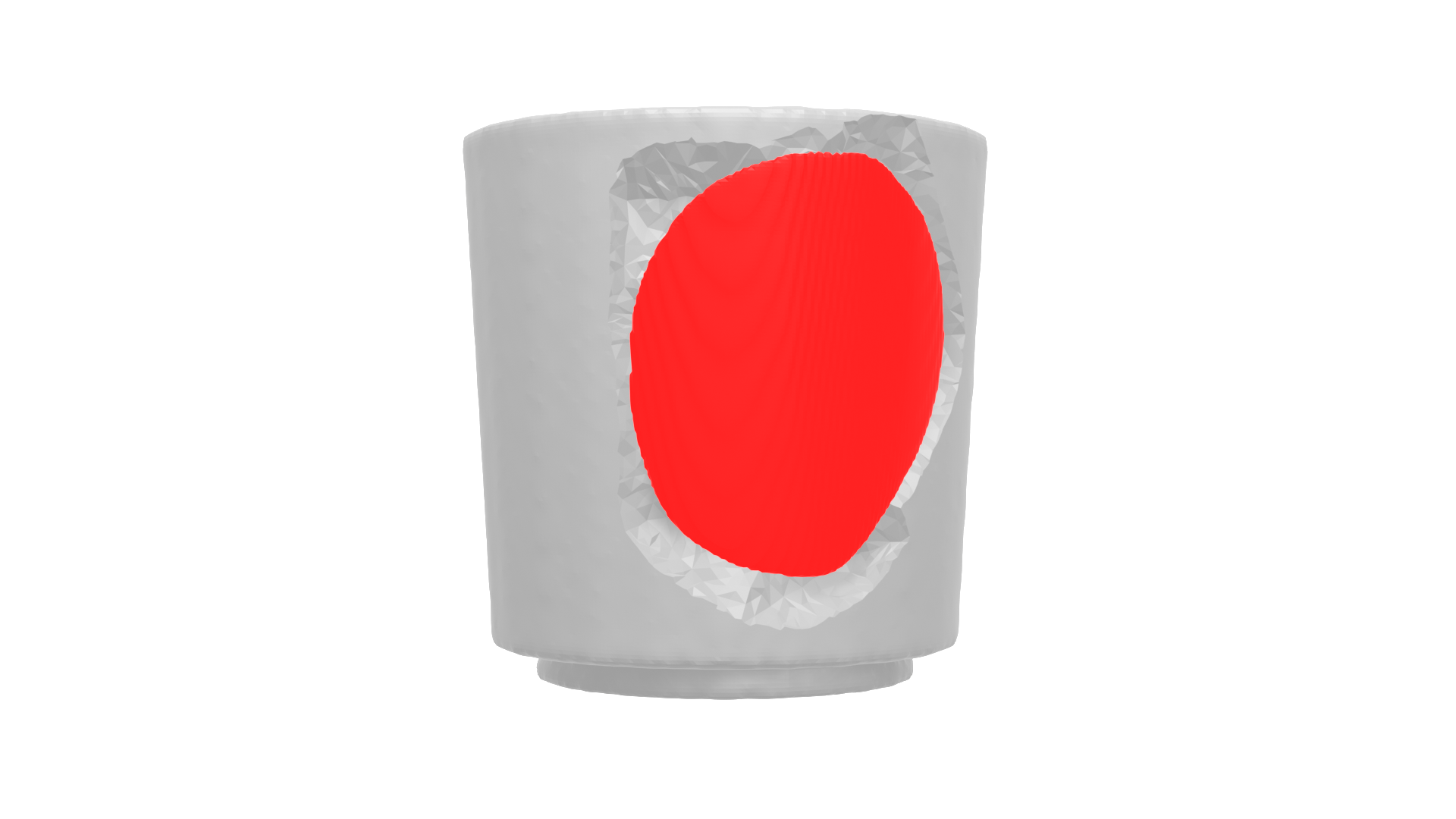}\\
    \small (b) DeepMend
  \end{tabular}~
  \begin{tabular}[b]{c}
    \includegraphics[trim={600 100 600 100},clip,width=.14\linewidth]{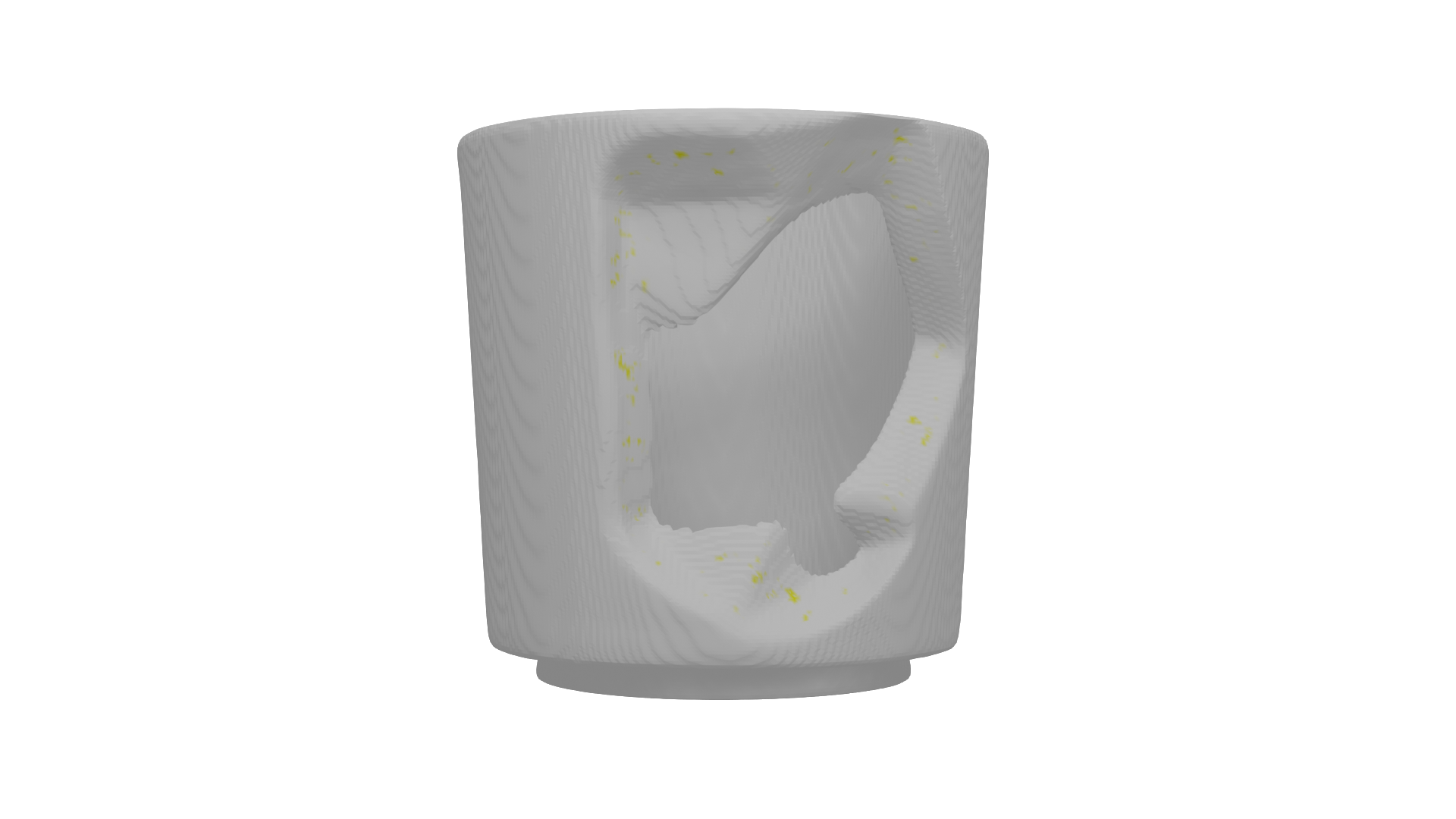} 
    \includegraphics[trim={700 220 700 290},clip,width=.14\linewidth]{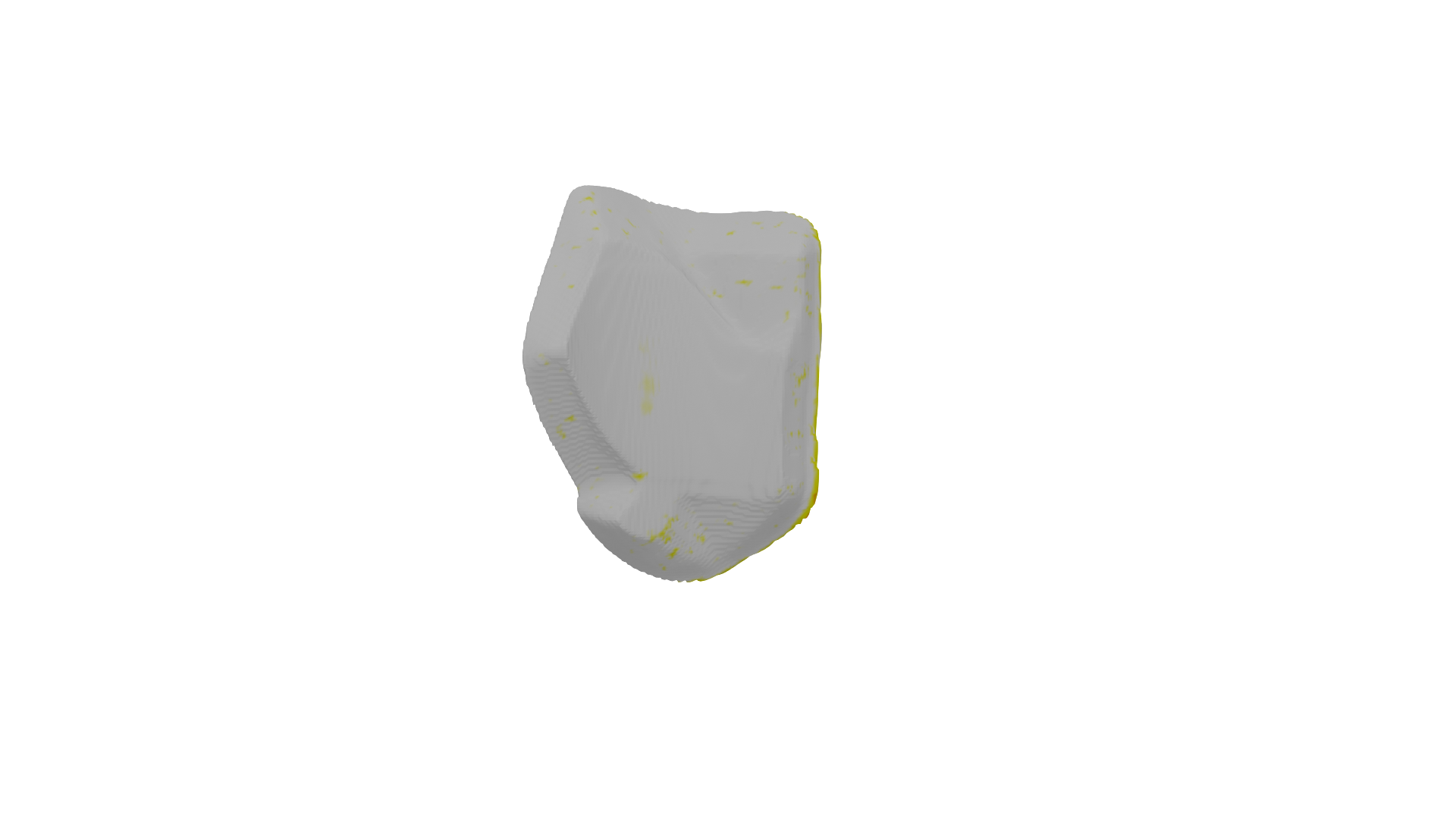}\\
    \includegraphics[trim={600 100 600 100},clip,width=.14\linewidth]{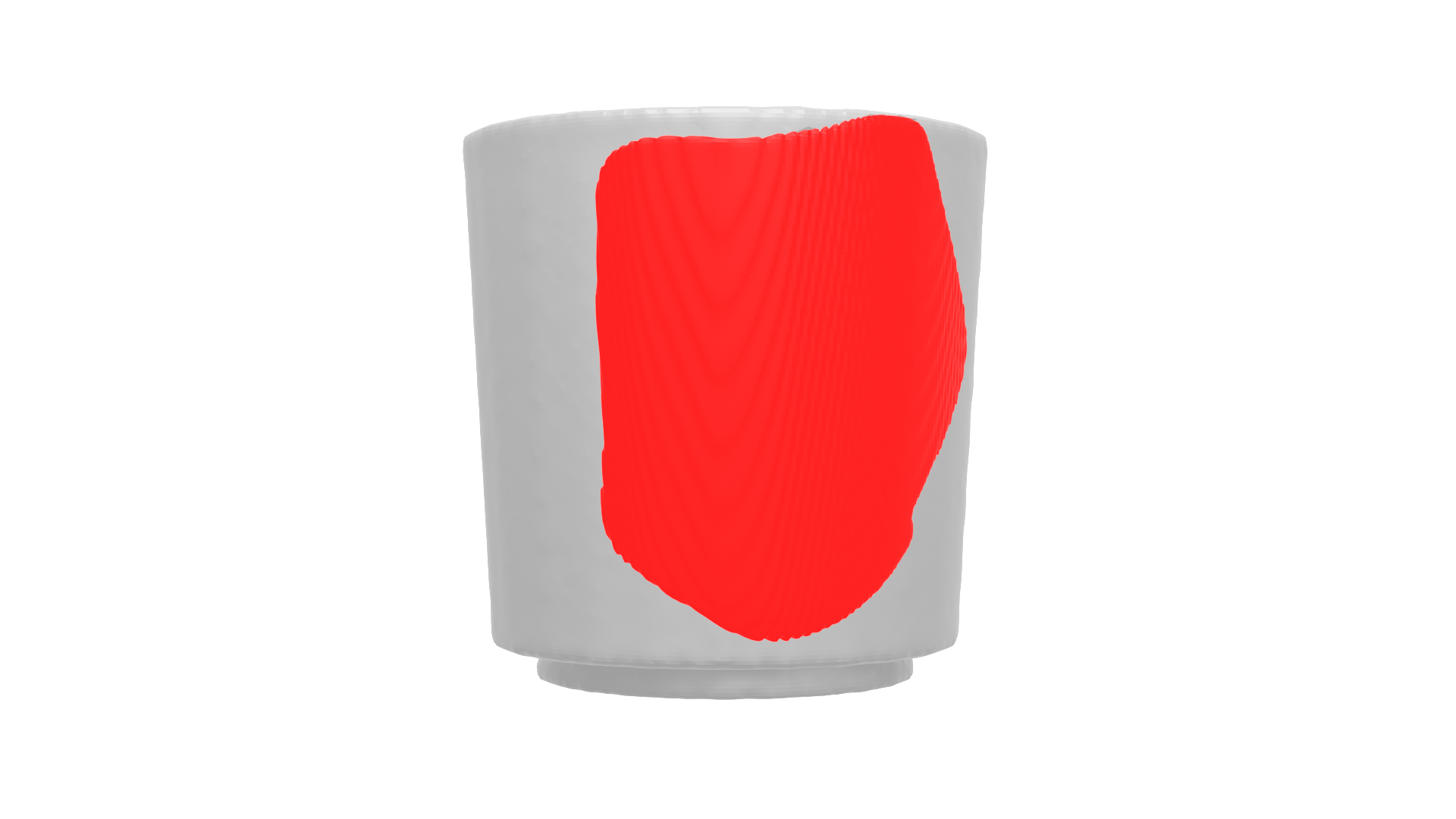}\\
    \small (c) Ours with \\
    test-time training
  \end{tabular}
  \includegraphics[width=0.5\textwidth]{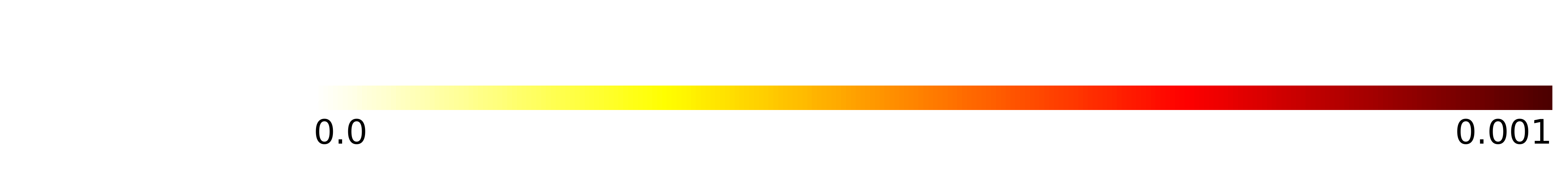}
  \caption{Overview about our method. While DeepMend gets a good rough estimation for the fractured (upper left) and restoration shape (upper right), we get via test-time training much sharper and more detailed shapes, especially w.r.t. the break surface. This results in restoration shapes that fit much better to the original input fractured shape (bottom). In particular, 3D printing as an application case will benefit from our approach.}
  \label{fig:teaser}
\end{figure*}

\input{chapters/01_introduction}
\input{chapters/02_Method}
\input{chapters/03_related_work}
\input{chapters/04_background}
\input{chapters/05_model}

\input{chapters/06_dataset}
\input{chapters/07_conclusion}

\noindent\textbf{Impact Statement. }This paper presents work whose goal is to advance the field of Machine Learning. There are many potential societal consequences of our work, none which we feel must be specifically highlighted here.

\noindent\textbf{Acknowledgements. } MM gracefully acknowledges funding of a Lamarr Fellowship of the Ministry for Culture and Science of the State of North Rhine-Westphalia. ZL was funded by a KI-Starter project of the Ministry for Culture and Science of the State of North Rhine-Westphalia during the majority of the project. The experiments in this paper were partially conducted using GPU resources of the
OMNI cluster at the University of Siegen, which we were kindly given access to by the HPC team.

\newpage
\bibliography{main}
\bibliographystyle{johd}

\end{document}

%% file: chapters/01_introduction.tex
\section{Introduction}\label{sec:introduction}


Partial objects are very common in shape analysis and pose special challenges when further processing the geometry. 
The partiality is often due to incomplete views when scanning an object which leaves holes in the reconstructed surface.
Many methods exist to complete these kinds of surfaces, both traditional and learning-based methods \cite{context-based_surface,shape_completion_anguelov,wu20153d}.
Another setting is the completion of partial volumetric shapes. 
In contrast to holes in surfaces, the boundaries of partiality are not obvious in this case and learned information about the space of shapes is crucial for obtaining good results. 
While such shapes can result from volumetric reconstruction algorithms \cite{slavcheva2017killing}, we focus on the case of \emph{shape restoration} in this paper. 
The assumption is that a broken object was scanned and the parts needed to complete it should be reconstructed (see \Cref{fig:teaser}). 
In addition to learning a distribution of complete shapes, it is necessary to reconstruct shapes that tightly align with the given input shape, for example to 3D print a fitting replacement part. 
This is, for example, important in medical applications from partial organ scans~\cite{gafencu2024shape} or for 3D printing parts for a broken object~\cite{deepmend2022}, e.g. for cultural heritage applications.

General shape completion methods focus on the appearance of the full reconstruction, which often looks good, but struggle with precision w.r.t. the input shape and details \cite{DeepSDF}.
We build our work upon \cite{deepmend2022} which tackles this problem and learns to complete fractured shapes in a certain class by separating the fractured part from the restoration part.
However, \cite{deepmend2022} still struggles to accurately align details, likely due to the complexity of learned classes whose details are hard to capture in a joint latent space. 
To overcome this issue we propose carefully designing the latent space and implementing test-time training to take the given geometric information of the input space into account directly. 
This allows us to produce restoration shapes that accurately represent the fractured area and produce a consistent joint shape, see \Cref{fig:teaser}.

\paragraph{Contributions. } We make the following contributions:
\begin{itemize}
    \item A new pipeline for generating accurate restoration shapes in the setting of volumetric shape completion. 
    \item An analysis and optimization of the network architecture for shape restoration proposed in \cite{deepmend2022}.
    \item The introduction of test-time training for shape restoration, which requires precise alignment and greatly benefits from this approach. 
    \item Code for Test-Time Training can be found here: \newline\url{https://github.com/mschopfkuester/shape_completion_ttt}
    \item Experiments on several datasets showing the advantages of the introduced changes. 
\end{itemize}

%% file: chapters/02_Method.tex
\section{Notation and Problem Description} \label{sec:notation}
In this section, we introduce the notation used throughout the paper and formalize the problem of shape restoration.

We assume that a \emph{fractured shape} $F\subset\mathbb{R}^3$ is given as input which is a partial version of a \emph{complete shape} $C\subset\mathbb{R}^3$. 
Therefore, it holds $F\subset C$. 
Our goal is to find the (with $F$ disjoint) \emph{restoration shape} $R\subset\mathbb{R}^3$ which completes $F$ when merged together, i.e. $C=F\, \dot{\cup}\, R$. 
See Figure~\ref{fig:f,r,c} for a visualization.
\begin{figure}[b!]
  \centering
  \begin{tabular}[b]{c}
    \includegraphics[trim={830 100 830 100},clip,width=.1\linewidth]{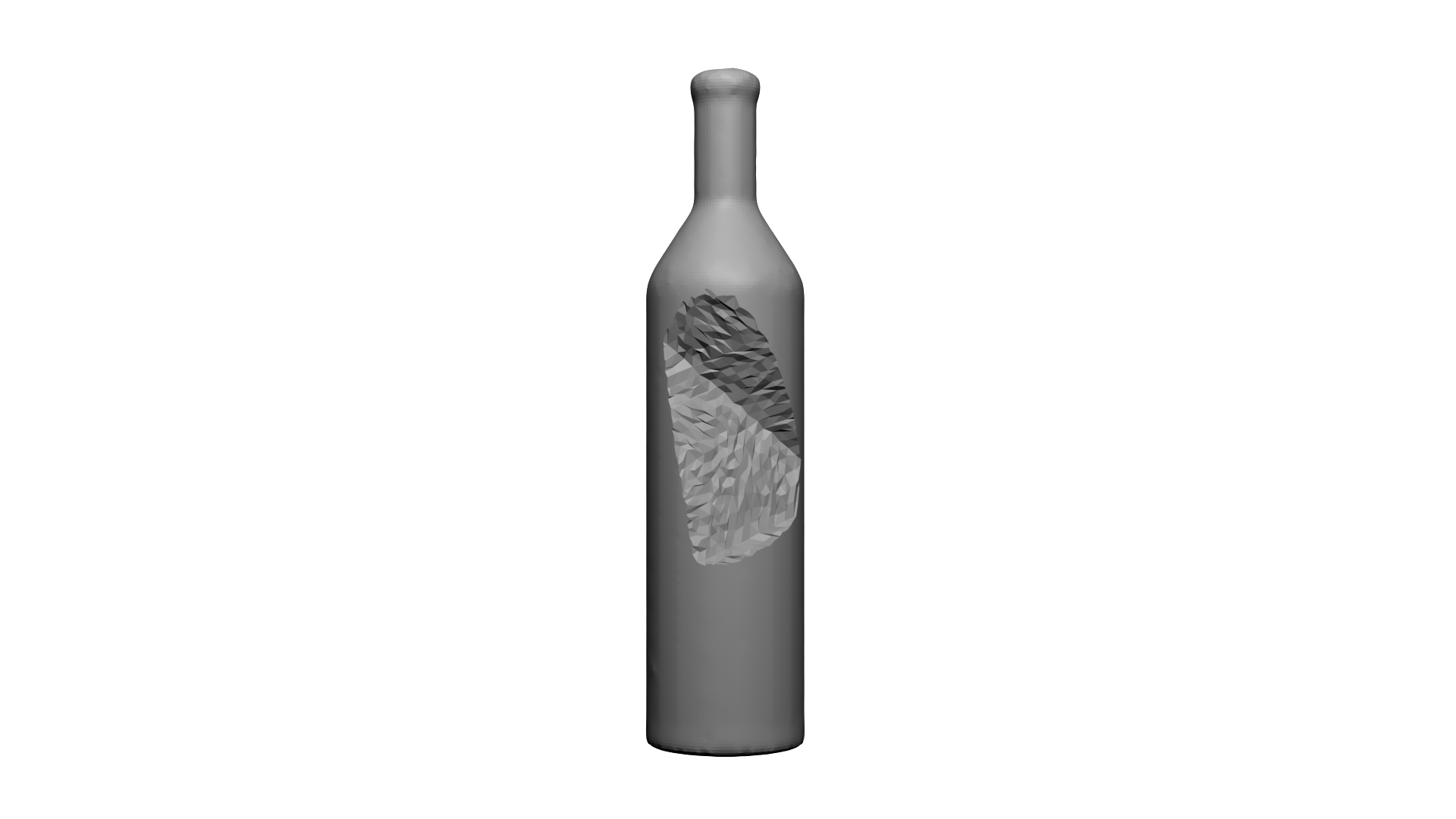} \\
    \small (a) \\ Fractured 
  \end{tabular}~
  \quad
  \begin{tabular}[b]{c}
    \includegraphics[trim={860 320 860 380},clip,width=.1\linewidth]{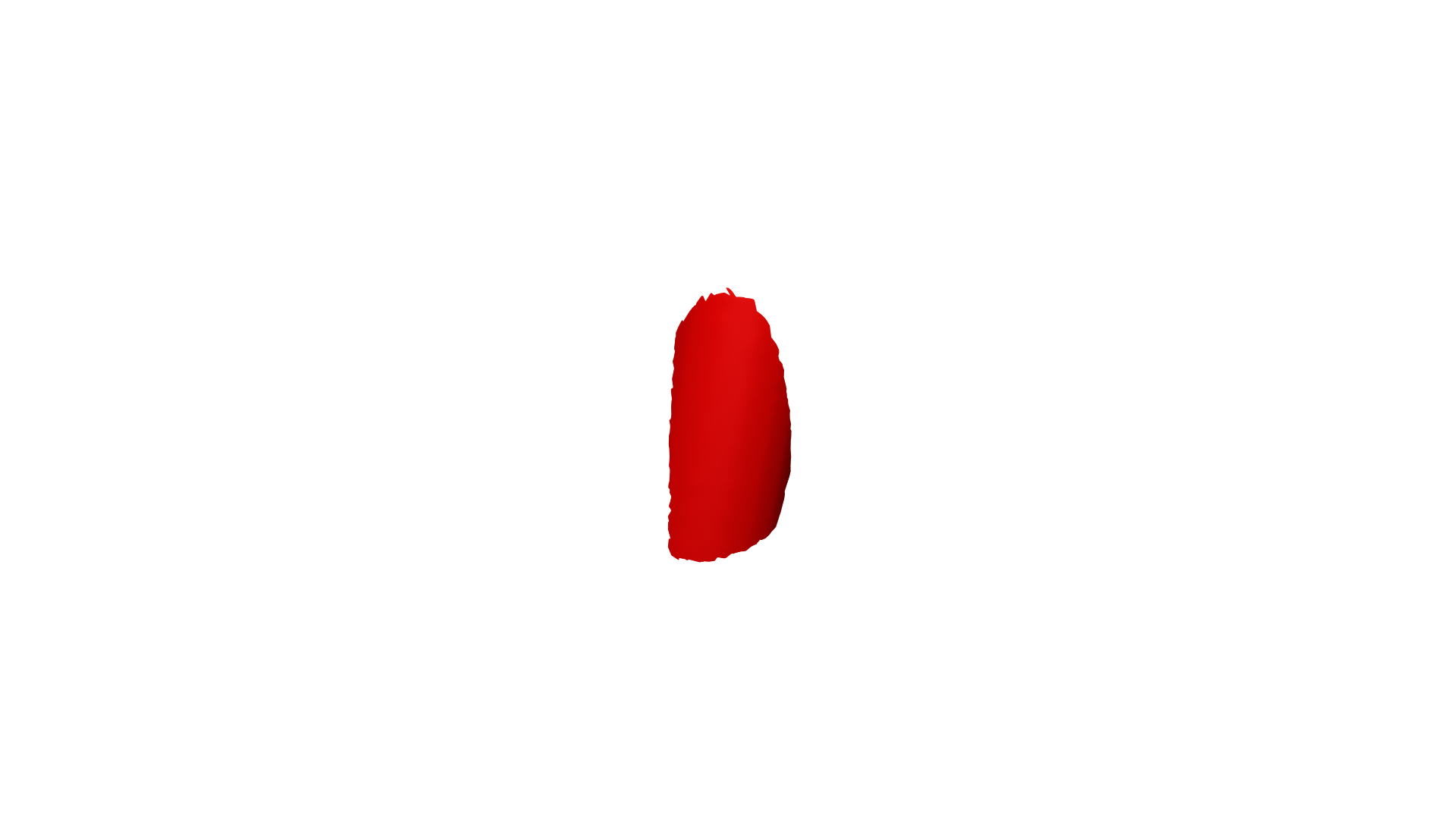} \\
    \includegraphics[trim={860 320 860 380},clip,width=.1\linewidth]{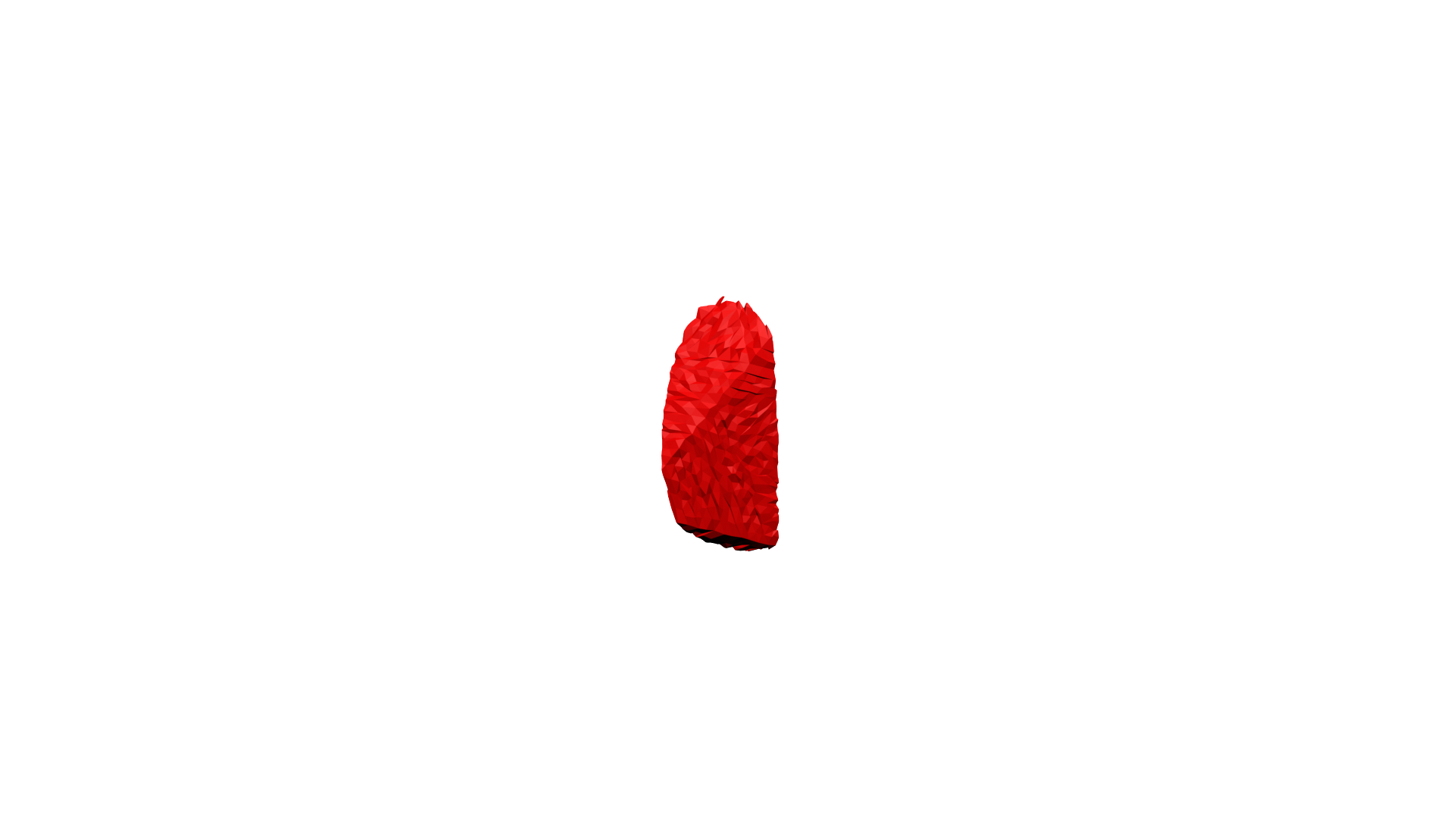} \\
    \small (b) \\ Restoration
    \end{tabular}~
    \quad
  \begin{tabular}[b]{c}
    \includegraphics[trim={830 100 830 100},clip,width=.1\linewidth]{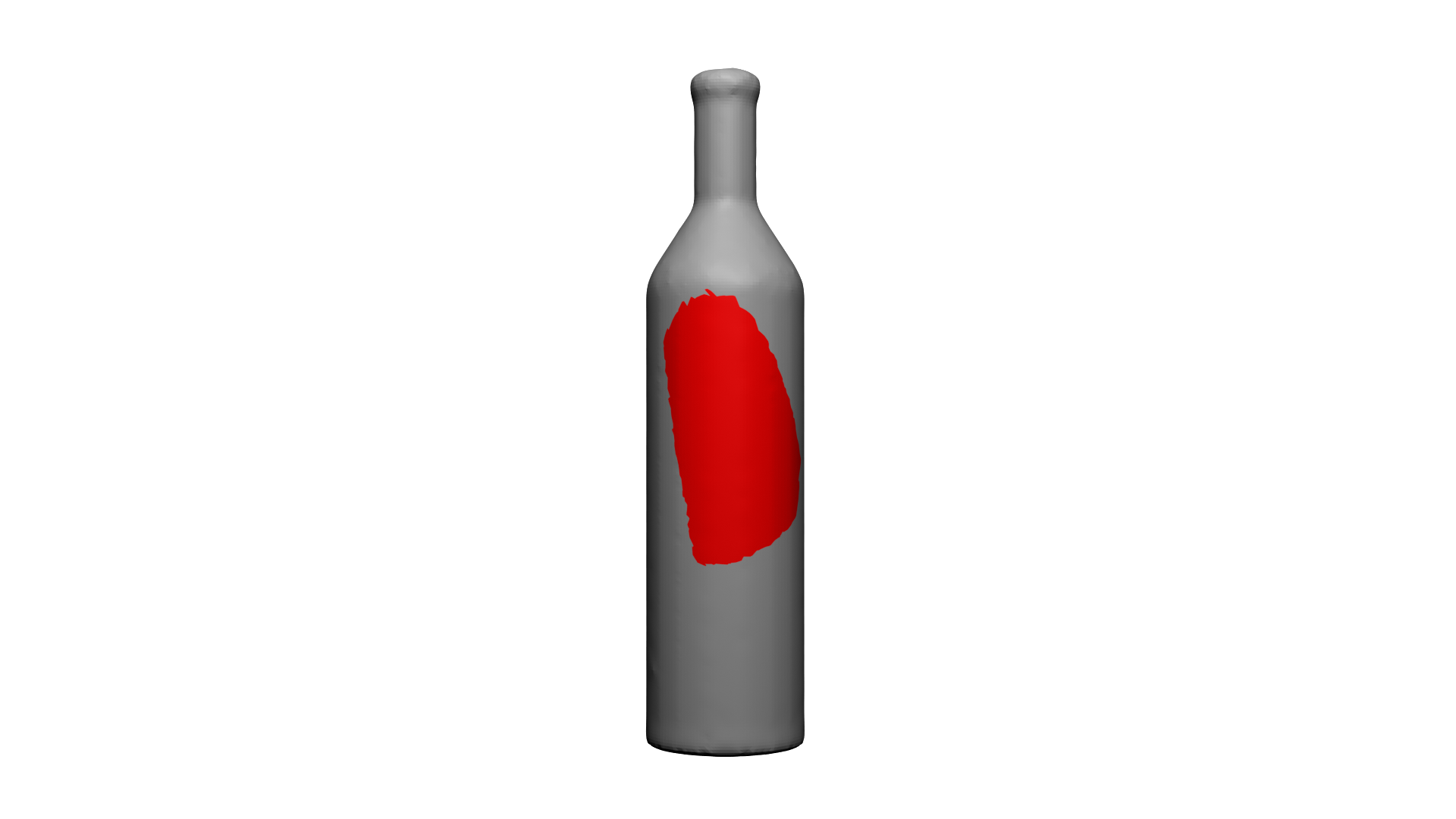} \\
    \small (c) \\ Complete
  \end{tabular}
  \caption{Visualization of the problem setting: Given a fractured shape $F$ (a), we want to predict the missing restoration shape $R$ (b) such that we get the complete shape $C=F\cup R$ (c).}
  \label{fig:f,r,c}
\end{figure}

We do not predict the complete shape directly, but the fractured and restoration shape separately. 
This ensures we can learn structural details about the relationship between the fractured and restoration part.
We need a few considerations to reformulate the problem accordingly. Therefore we  introduce a so called \emph{break set} $B \subset \mathbb{R}^3$ such that both, the fractured shape $F$ and restoration shape $R$, can be described as the intersection of the predicted complete shape and the break set (or its complement), i.e. 
\begin{align}
    F=C \cap B \text{ and } R=C \cap B^C.
\end{align}
%
This allows to model $F$ and $R$ separately (and pairwise disjunctive) from each other but also guarantees an (inner) relationship between both shapes.

We describe the surface of all shapes by occupancy functions $o_S(x): \mathbb{R}^3 \to \{ 0, 1 \}, S \in \{ F, R, C, B \}$, such that for a certain point $x\in\mathbb{R}^3$ it holds that
\begin{align}
    o_S(x)=\begin{cases}
        1,& x\text{ is inside the shape } S,\\
        0,& \text{other}.
    \end{cases}
\end{align}
In the binary formulation, $o_F$ and $o_R$ can be easily rewritten in the following from $o_C$ and $o_B$:
\begin{align}
    o_F(x)&=o_C(x)\cdot o_B(x), \label{eq:oF}\\     
    o_R(x)&=o_C(x)\cdot(1-o_B(x)) \label{eq:oR}.
\end{align}
This allows an efficient on-the-fly conversion between all entities. 

%% file: chapters/03_related_work.tex
\section{Related Work}\label{sec:related_work}

In this section, we provide an overview of general 3D generative models as well as shape completion approaches. 

\subsection{3D Generative Models}

Generating 3D geometry has been a widely studied field~\cite{dai2017completion,luo2021probpcs,qiu2023richdreamer} which has greatly benefited from the advances in implicit representations due to their flexibility~\cite{DeepSDF,erkoc2023hyperdiff}.
These methods are fundamentally different from mesh-based methods like deformation models~\cite{loiseau2021linearmodels} or parametric shape models~\cite{zuffi2018smal} which rely on a fixed template or topology and, thus, are limited in the shapes they can represent. 
Combining both can lead to great results but might suffer from inconsistencies between them~\cite{poursaeed2020coupling,Mehta2022levelset}.
Generative models based on neural fields can handle different classes and be conditioned with arbitrary modalities, for example, images~\cite{gao2022get3d,liu2023one2345}, text~\cite{qiu2023richdreamer,lin2023magic3d}, or latent code manipulation~\cite{DeepSDF,zeng2022lion,hu2024topology}.
A special case of generative modeling is \textit{shape completion}, which takes an incomplete or broken shape as input and aims to generate the full object.

\subsection{Shape Completion}

The partial input can either be an incomplete point cloud 
or a partial volumetric shape for which semantic information is necessary for completion. 
A partial point cloud or mesh has identifiable holes which can be closed without any information about the object class, for example by Poisson reconstruction which generates water-tight surfaces out of any oriented point cloud~\cite{Kazhdan:2006:poisson}, by filling the hole with structurally fitting patches~\cite{hanocka2020point2mesh} or using learned class features~\cite{chibane20ifnet}.
Learning class-based priors for point cloud completion allowing to complete more severe degradation is also possible~\cite{bo2022patchrd,zhu2023svdformer,cui2024neusdfusion}.

In a volumetric representation, the boundaries of holes are harder to identify, and semantic information about the properties of the full shapes in this class is necessary. 
One of the first works in this direction applied convolutions on voxel grids in a coarse-to-fine manner to complete incomplete depth fusions~\cite{dai2017completion}.
However, voxel methods require a lot of memory to represent fine details.
%
With \cite{breakingbad} and \cite{fantasticbreaks} two datasets with complex fractured shapes were published recently.
%
A huge step towards more efficient detail generation was made in DeepSDF~\cite{DeepSDF} which learns to predict a signed distance function (SDF) with a neural network. 
DeepSDF can do both, sample new instances from a class as well as complete partial shapes.
However, a strong class prior learned by the network can lead to semantically meaningful but not well-aligned completed shapes.

In applications where a tight fit to the input is necessary, for example, 3D printing replacement parts of broken objects (\emph{shape restoration}), this is a problem. 
This can be avoided by requiring additional information, like an image of the complete shape~\cite{galvis2024scdiff}.
Another solution was proposed by DeepMend~\cite{deepmend2022} and DeepJoin~\cite{lamb2022deepjoin} in explicitly modeling the fractured region and the fit of the generated part to the fracture.
Due to the more general setup, where only the occupancy of the shapes is needed, we focus on~\cite{deepmend2022}.
We provide a more detailed description of DeepMend in Section~\ref{sub:bg:deepmend}.
But like many generative methods, DeepMend suffers from over-smoothing behavior from having a single network for many objects, as we can see in our numerical experiments.
To that end, we propose using test-time training for shape restoration which can adjust the network weights to the input and ensure that a tight fit is possible. 

\subsection{Test-Time Training}

The power of learning methods normally comes from huge generalization capabilities based on the training data. 
However, examples that stray from the training distribution might only be captured sufficiently but not perfectly -- often seen in slightly too smooth test results.  
Finetuning the network (or parts of it) during interference can prevent this in applications, where it is possible to do self-supervised adaption and where inference time is not critical. 
This so-called test-time training was proposed in \cite{sun2020ttt} for generalizing under distribution shifts and has been applied in different applications like robust classification~\cite{gandelsman2022ttt} and sketch-based image retrieval~\cite{sain2022sketch}. 
To the best of our knowledge, this work is the first to use test-time training to improve the geometric consistency in learning-based shape completion approaches by overfitting the network weights to match the (fractured) input shape.

%% file: chapters/04_background.tex
\begin{figure}[t]
    \centering
    \includegraphics[width=0.9\linewidth]{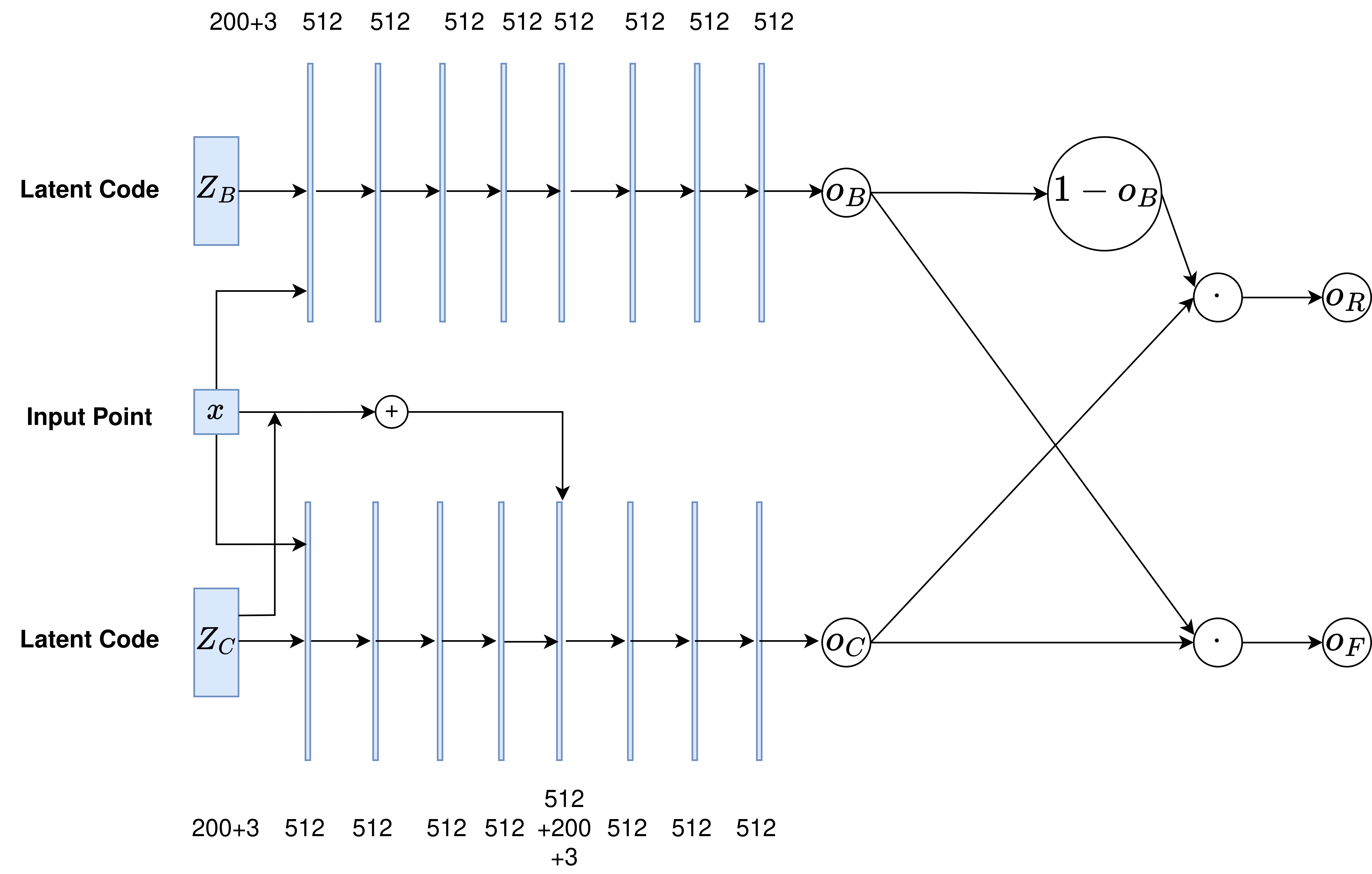}
    \caption{Network architecture of~\cite{deepmend2022}. The architecture is separated into two parts which predict the occupancy of the complete shape $o_C$ and the break set $o_B$ respectively. The input for both parts is the point coordinate $x\in\mathbb{R}^3$ and a latent code describing the geometry. Via  \Cref{eq:oF} and \Cref{eq:oR} we can calculate the occupancies of the fractured shape $o_F$ and the restoration shape $o_R.$ We define the skip connection with $\oplus$ and the multiplication of the outputs of the two networks with $\odot$.}
    \label{fig:model}
\end{figure}

\section{Background}\label{sec:background}

In this section, we introduce the most relevant previous work, DeepMend~\cite{deepmend2022}, which tackles the shape restoration problem and on whose network architecture our work is built, in more detail.

\subsection{DeepMend} \label{sub:bg:deepmend}

DeepMend \cite{deepmend2022} learns to predict the occupancy functions $o_C$ and $o_B$, as defined in \Cref{sec:notation}, via neural networks and thus learns a representation for the relationship between the fractured and restoration shape that generalizes well to new class instances.
The network architecture is an auto-decoder based on DeepSDF~\cite{DeepSDF} which has as input parameters a latent code and a point $x\in\mathbb{R}^3$ and predicts the corresponding occupancy values $o_C(x)$ and $o_B(x)$. The latent code is optimized directly on every instance without a trained encoder with a decoder-only framework, see DeepSDF for details. 
%

%
To predict the occupancy of the complete shape $o_C(x)$, \cite{deepmend2022} uses a network $f_{\theta_1}$ consisting of a $8$-layer MLP with a skip connection after the fourth layer and a $128$-dimensional latent code $z_C$ as input. 
A similar network $g_ {\theta_2}$ is used to predict the break set but without a skip connection and a smaller latent code $z_B$ as input. 
The exact architectures can be found in \cite{deepmend2022}. See also Figure \ref{fig:model} for an overview. 




\begin{figure*}[t]
    \centering
    \includegraphics[width=0.9\textwidth]{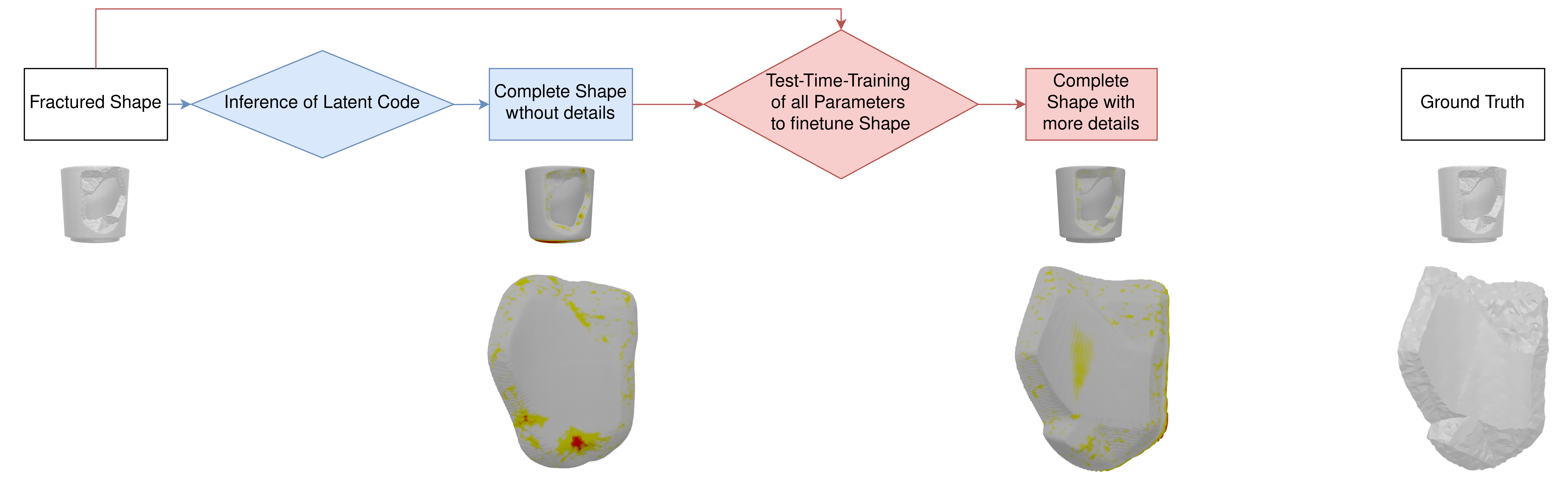}
    \includegraphics[width=0.5\textwidth]{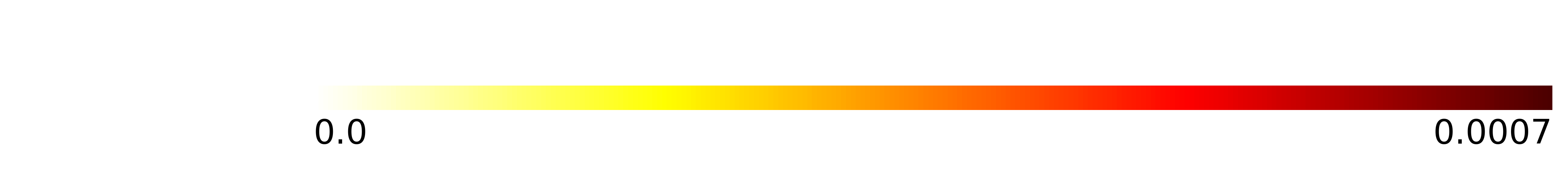}
    \caption{Pipeline of our method with test-time training. After only optimizing the latent code to get a rough prediction of the restoration shape (blue, this is the pipeline of DeepMend~\cite{deepmend2022}), we use the predicted complete shape as well as the input fractured shape to finetune all network parameters and therefore get a more detailed restoration shape (red, our addition).}
    \label{fig:enter-label}
\end{figure*}

\paragraph*{Training and Inference}
For model training \cite{deepmend2022} considers the binary cross-entropy loss (BCE) $\mathcal{L}_S$ between the true target $\hat S$ and the corresponding prediction $S$ for each of the four sets $F,R,C,B$:
\begin{align}
\mathcal{L}_S=\sum_{x}BCE(o_S(x,z_B,z_C,\theta_1,\theta_2),o_{\hat S}(x)). \label{eq:bce}
\end{align}
The specific terms can be derived from \Cref{eq:oF} and \Cref{eq:oR}.
Each object class is trained separately and until the chamfer distance error on the validation set is minimal. 
Dropout is used on all hidden layers.
During training, the network parameters $\theta_1$ and $\theta_2$ are jointly optimized with the instance-specific latent codes $z_B$ and $z_C$:
\begin{align}
    \{\hat z_B^j\},\{\hat z_C^j\},\hat \theta_1,\hat \theta_2&=\argmin_{ \{z_B^j\},\{z_C^j\},\theta_1,\theta_2}\mathcal{L}_{\text{train}} \notag \\
    &=\argmin_{ \{z_B^j\},\{z_C^j\},\theta_1,\theta_2}\mathcal{L}_C+\mathcal{L}_B+\mathcal{L}_F+\mathcal{L}_R,
\end{align}
where $\mathcal{L}_C,\mathcal{L}_B,\mathcal{L}_F,\mathcal{L}_R$ are the respective binary cross-entropy losses for the complete shape, break set, fractured, and reconstruction shape for the training examples $j=1,\dots,n$. The losses are taken from \Cref{eq:bce}.
During inference, only the latent codes are optimized and the fractured shape is given as input, such that we calculate
\begin{align}
    \hat z_B,\hat z_C &= \argmin_{z_B,z_C} \mathcal{L}_F+\mathcal{L}_{reg}
\end{align}
for each test object separately. Here $\mathcal{L}_{reg}$ is a combination of different penalty terms to ensure that the predicted restoration shape is not empty and near the fractured shape. The final shape is then inferred as $f_{\theta_1}(\hat z_C).$
For more information see \cite{deepmend2022}.

%% file: chapters/05_model.tex
\section{3D Shape Restoration}\label{sec:method}

In this section, we introduce our method to complete fractured shapes based on test-time training.
We build upon the fact that while predicting the rough geometry of the whole shape is important, it is equally important to fit the prediction \emph{exactly} to the partial input to not create artifacts around the break points and, for example, allow a properly fitting spare part to be printed, see Figure~\ref{fig:teaser}. 
To that end, we apply test-time training on the network weights to allow a tight fit to the input geometry, see Section~\ref{subsec:ttt}.
Additionally, we perform an analysis of the network architecture and loss functions used in \cite{deepmend2022} to increase the performance, see Section~\ref{subsec:arch}.

\subsection{Test-Time Training} \label{subsec:ttt}
%

%
%
%
%
%
During test-time training, the model's weights are finetuned using input data during inference \cite{sun2020ttt}. Models trained on complex classes often overlook finer details, resulting in smoothed edges and missing details, as seen in \cite{deepmend2022}. By incorporating the detailed geometric information from the input fractured shape into test-time training, we achieve better geometric consistency, as demonstrated in our experiments.

Since the full shape is unknown during inference, the training procedure must be adapted to a self-supervised loss function. 
To that end, we predict the complete shape $C$ using the normally trained network, conditioned on the input fractured shape $\hat F$.
%
%
Then, we get the corresponding restoration shape $\hat R: = C\setminus \hat F$ and use $\hat F$ as well as $\hat R$ to finetune the network with two loss functions, preserving their geometry:
%
\begin{align}
    \hat z_B,\hat z_C,\hat \theta_1,\hat \theta_2&=\argmin_{ z_B,z_C,\theta_1,\theta_2}\mathcal{L}_{\text{TTT}} \notag \\
    &=\argmin_{ z_B,z_C,\theta_1,\theta_2}\mathcal{L}_{F}+\alpha \cdot \mathcal{L}_R .
\end{align}
In ${L}_{F}$ the geometry of the input shape is aligned and incentives the network to tightly fit to the known details.
%
%
To prevent complete overfitting onto $\hat F$ and an empty restoration shape, the class information from the pretrained network is preserved implicitly in $\mathcal{L}_R$.
%
%
We choose $\alpha=0.1$ in all experiments.  
The network is trained for $3000$ epochs during test-time training.

\subsection{Network Architecture} \label{subsec:arch}

In addition to the test-time training, we conducted an analysis of the network architecture used in \cite{deepmend2022} and propose two changes that increase the performance:
\begin{enumerate}
    \item First, we increase the dimensionality of the latent codes $z_C$ and $z_B$. To properly capture geometric details a larger latent space is beneficial, especially if the details can be fine-tuned during test-time training. We see optimal results with $\dim z_C=\dim z_B=200$.
    \item We change the design of the skip connection. In DeepMends architecture the skip connection replaces part of the neurons of the hidden layer instead of concatenating them. This architecture design reduces the usable latent dimensionality as well as the expressivity of the network. Instead we concatenate the information such that the hidden dimension is $512+\dim z+\dim x=715$ instead of $512$.
\end{enumerate}
Overall, these changes increase the number of parameters of our network design, which makes it harder to train (see \Cref{table: 10per} DeepMend vs. Ours w/o TTT).
However, in combination with the test-time training the additional degrees of freedom allow us to represent the input data a lot more accurately.

%% file: chapters/06_dataset.tex
\section{Experiments} \label{sec:experiments}

\subsection{Implementation Details}
Experiments are run in Pytorch 2.0.1 with CUDA version 11.7. All experiments are trained and evaluated on one NVIDIA GeForce RTX 3090 and AMD Ryzen 9 5900. Training one model took 8 hours, inference of the latent code 20 minutes and test-time training additional 30 minutes. The inference time for \cite{deepmend2022} was 45 minutes. Our model has 3,206,652 parameters and needs $2.03\cdot 10^{9}$ additions and multiplications for one forward pass while \cite{deepmend2022} has 2,931,964 parameters and needs $1.896\cdot 10^{9}$ additions and multiplications for one forward pass. We used Adam optimization with a learning rate of $5\cdot 10^{-4}$ for the network parameters and $10^{-3}$ for both latent codes. The choice of the hyperparameters for inference are taken over from \cite{deepmend2022} and can be found there. 
For all experiments, we use Marching Cubes \cite{MarchingCubes} to generate meshes from the implicit representation. 

\begin{table*}[t]
\centering
\begin{tabular}{l||r|r|r|r|r|r|r|r|}
 & \multicolumn{2}{c|}{DeepSDF}  & \multicolumn{2}{c|}{DeepMend} &  \multicolumn{2}{c|}{Ours w/o TTT}& \multicolumn{2}{c|}{Ours w/ TTT}  \\ 
 & Mean & Median & Mean & Median & Mean & Median & Mean & Median  \\ \hline
 airplanes& 2.36 &0.61  & 6.37 &1.92  & 4.67 &0.98 & \textbf{1.78} &\textbf{0.49}  \\
 bottles& \textbf{4.37} &1.43  & 5.94 & 0.45 & 6.56 & \textbf{0.35} & 5.09 & \textbf{0.35} \\
 cars& 5.18 & 3.07 & 4.70 &2.60  &  5.21 & 2.82 &\textbf{3.15}& \textbf{1.14} \\
 chairs& \textbf{8.36} & 5.31 &  20.08& 9.50 &21.69  &12.57  & 11.13 &\textbf{1.5}  \\
 jars& \textbf{13.65} &5.93  & 37.92 & 6.65 & 28.52 &6.95  & 16.40 & \textbf{1.50} \\
 mugs& 4.26 &\textbf{1.00} & 5.27 &3.02  & 5.34 & 1.02 & \textbf{3.24} & 1.21 \\
 sofas& \textbf{4.29} &2.60  & 8.10 & 3.57 &  9.39& 4.13 & 5.60 &\textbf{1.73} \\
 tables& \textbf{8.41} &\textbf{3.35}  &28.23  &9.59  &  26.74& 12.28 & 17.25 & 4.98 \\ \hline
 Mean& \textbf{6.36} & 2.91 &14.58& 4.65 & 13.52 &5.14  &7.96  &\textbf{1.43}
\end{tabular}
\caption{Chamfer Distance ($\cdot 10^{-4}$) for the ShapeNet Dataset~\cite{shapenet2015} on Dataset 1 in which the complete shapes have $5-20\%$ percent of their volume removed. Lower is better. The best value within each class is set \textbf{bold}.}
\label{table: 10per}
\end{table*}

\begin{table*}[ht!]
\centering
\begin{tabular}{l||r|r|r|r|r|r|r|r|}
 & \multicolumn{2}{c|}{DeepSDF}  & \multicolumn{2}{c|}{DeepMend} &  \multicolumn{2}{c|}{Ours w/o TTT}& \multicolumn{2}{c|}{Ours w/ TTT}  \\ 
 & Mean & Median & Mean & Median & Mean & Median & Mean & Median  \\ \hline
 airplanes& 12.23&4.29    &8.35 &1.49  &5.66  &\textbf{1.1 }&\textbf{4.18} &  1.41    \\
 bottles& 15.79&\textbf{8.07}   &\textbf{11.84}  &9.41  & 13.85& 9.37 & 13.7 & 9.04\\
 chairs&49.97 &29.88    &\textbf{22.34}  &\textbf{10.1}  &24.23 &16.27  & 24.67 &12.30  \\
 jars& 74.86& 69.97   &35.29  &22.82 & 38.27 &  25.51 & \textbf{32.63} &\textbf{17.04} \\
 mugs&42.13 &11.18   &\textbf{15.09}  &6.2  & 16.19 &5.06  & 15.23  & \textbf{4.71} \\
 sofas& 27.10& 17.36  &11.44  &7.6  & 11.31 &6.0  &\textbf{9.84} &\textbf{4.78}\\
 tables& 43.29& 28.52   &27.63  &20.57  &21.89 &13.08  &\textbf{18.79}  & \textbf{7.62}\\ \hline
 Mean& 33.62&24.18    & 18.85 & 11.17  &  18.77 &10.21 & \textbf{17.00}&\textbf{8.13}
\end{tabular}
\caption{Chamfer Distance ($\cdot 10^{-4}$) for the ShapeNet Dataset~\cite{shapenet2015} on Dataset 2 in which the complete shapes have $45-55\%$ of their volume removed. Lower is better. The best value within each class is set \textbf{bold}.}
\label{table: 50per}
\end{table*}
\begin{figure*}[t!]
    \centering
    \includegraphics[width=0.87\linewidth]{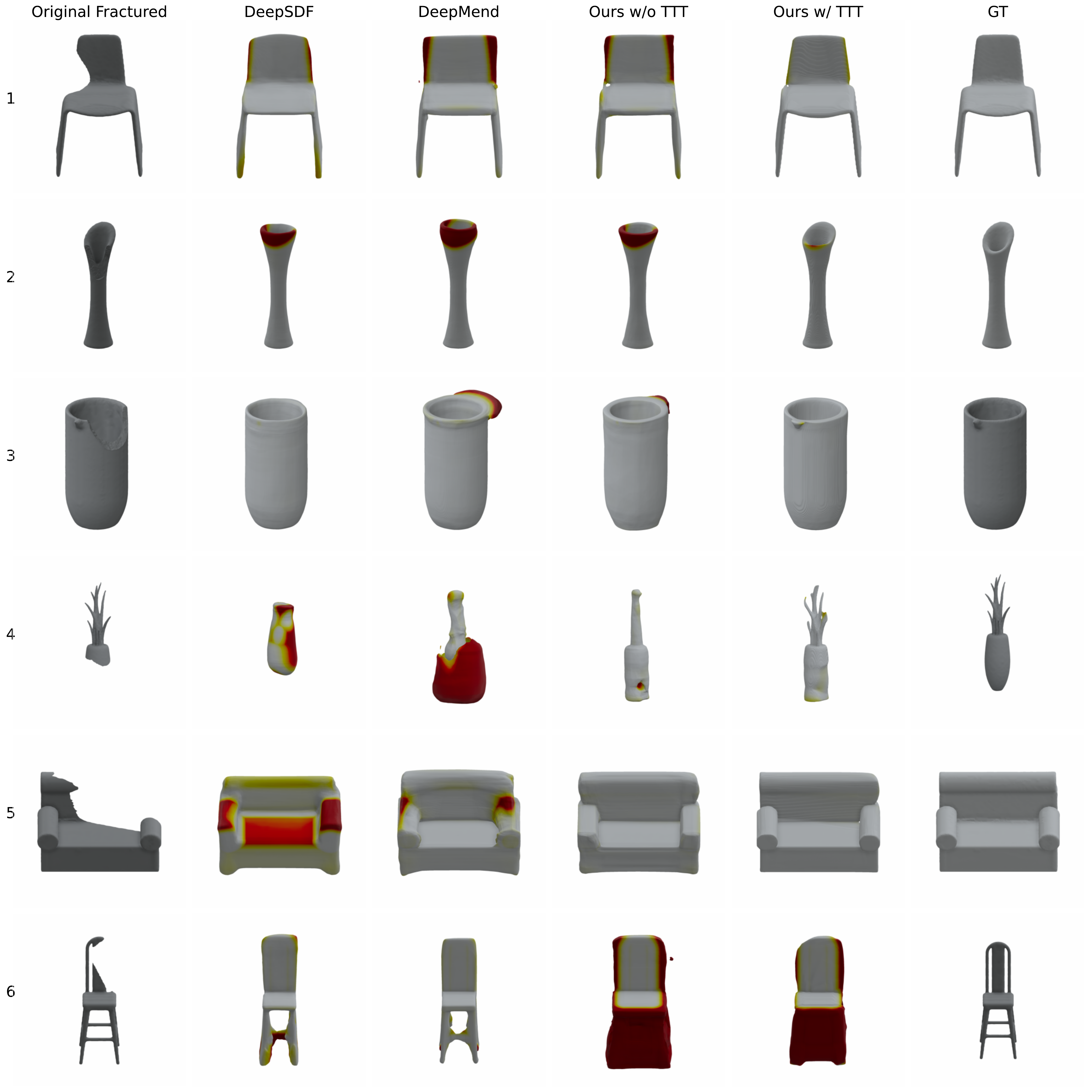}
    \includegraphics[width=0.5\textwidth]{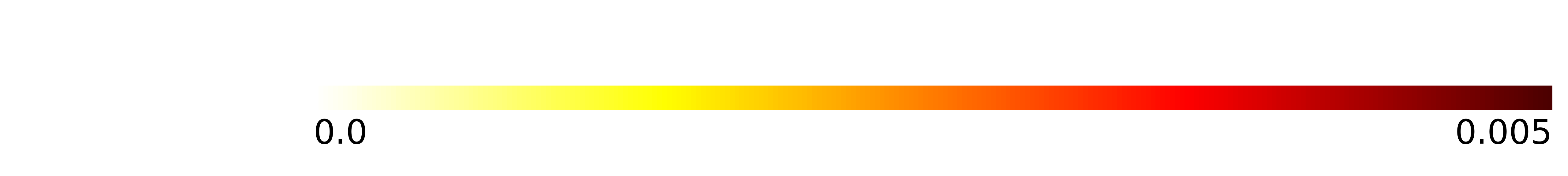}
    \caption{Qualitative examples of the different methods. Rows 1-3 and 6 are taken from Dataset 1, Rows 4-5 are from Dataset 2. Even though the Chamfer distance does not change much, the ability of our method to fit the fractured shape well, does visually a huge difference.}
    \label{fig: examples}
\end{figure*}

\subsection{Dataset}
We evaluate our method on ShapeNet~\cite{shapenet2015} which contains over 50 000 real-world scanned 3D objects of different classes. We train each object class separately and use a $70/15/15$ train/validation/test-split and $240$ objects per class.

For preprocessing we fracture all objects using the approach of \cite{CreateFracture2021}. 
For different levels of complexity we create two different settings in which the percentage of removed volume varies. 
In \textbf{Dataset 1} the complete shapes have $5-20\%$ of their volume removed, and in \textbf{Dataset 2} it is $45-55\%$. 
We sample random points within the unit cube, calculate the signed distance function to determine if each point is inside or outside the mesh, and use this information to compute the occupancy of the the shape at the given points.

\subsection{Metric}
As evaluation metric we use the Chamfer distance (CD) between the ground-truth and predicted complete shape. It is defined as follows:

\begin{align}
    d_{\text{CD}}(X,Y)= & \frac{1}{\abs{X}}\sum_{x\in X} \min_{y\in Y} \norm{x-y}^2_2 \notag  \\
    +& \frac{1}{\abs{Y}}\sum_{y\in Y} \min_{x\in X} \norm{x-y}^2_2,\quad X,Y\subset \mathbb{R}^3.
\end{align}

\subsection{Comparison}

We compare ourselves to two main competitors. 
First, the classic method of DeepSDF~\cite{DeepSDF} which predicts the full SDF from a given partial shape. 
Second, the approach of DeepMend~\cite{deepmend2022} which, while based on DeepSDF, proposed the separation into fractured and restoration shape. 
All tables also contain "Ours w/o TTT" which is our adapted network architecture of DeepMend (see \Cref{subsec:arch}) but without test-time training during inference. 
Notice that test-time training cannot be applied to DeepSDF directly as fine-tuning the complete shape to the input directly would lead to a complete overfitting of the input.  

Our test-time training does add additional compute during inference. 
To make the comparison fairer, we reduce the number of training iterations for our method by the amount we use for test-time training. 
This means for every single example all methods had the same computing power at their disposal.

\subsection{Results}

\paragraph{Dataset 1, Low Partiality}

\begin{figure*}[t]
    \pgfplotsset{/pgfplots/group/.cd,
        horizontal sep=2.5cm,
        vertical sep=2cm
    }
    \centering
    \input{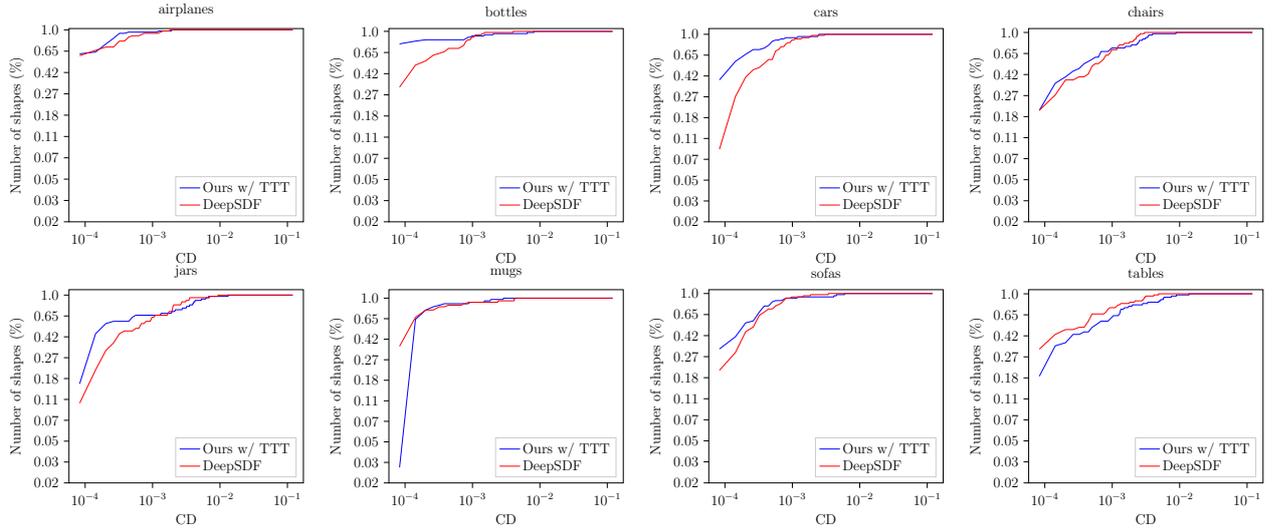}
    \caption{Logarithmic cumulative error curves (higher is better) of the chamfer distance  (CD) for the different classes of Dataset 1. We compare our approach with test-time training against DeepSDF. We see that our model produces for more objects a smaller Chamfer distance, even though our model has more outliers.}
    \label{fig: cumulative error}
\end{figure*}

Table~\ref{table: 10per} summarizes the errors on Dataset 1, which contains examples in which $5-20\%$ of the shapes are removed, for DeepSDF, DeepMend and our model with and without test-time training. 
For each class, the mean and median of the Chamfer distance over all test objects within one class is depicted. 
We can observe that the small adjustments of the network architecture (dimension of latent code, skip connection, see \Cref{subsec:arch}) led to improvements over DeepMend in most classes. 
Using test-time training, we improve the CD of our model in each category and overall by 41 percent (by even more for the median distance). 
Even though DeepSDF beats our model w.r.t. the mean most of the time, our median is lower in 6 of 8 categories. 
For a better understanding of the different behavior for mean and  median, we refer to cumulative error curves in Figure~\ref{fig: cumulative error}. 
We see that our model produces more objects with a lower error but also a few bad outlier which influences the mean a lot.
An example for this behaviour can be found in Row 6 of Figure \ref{fig: examples}.

\paragraph{Dataset 2, High Partiality} The second dataset presents a bigger challenge as the removed volume is increased to $45-55\%$. 
The results in Table~\ref{table: 50per} show that, as for Dataset 1, our method tends to perform slightly better than DeepMend before test-time training and significantly better after. 
However, test-time training only improves by $9.4\%$ which is due to the fact that the restoration part takes up around half of the shape on which the test-time training has less influence.
Interestingly, DeepSDF performs much worse in this category than with less partiality. 
Our assumption for this behaviour lies in the different frameworks of the methods: While DeepMend as well as our model is specifically designed to learn a meaningful relationship between the fractured and restoration shape, DeepSDF has to encode the entire class geometry in a single model which leads to complex problems under harsh partiality. 
An indicator for this is that DeepSDF does perform quite well on the bottles class, which has the least intraclass variations, and is, thus, easier to capture in a single model. 

\paragraph{Qualitative Results} We show some qualitative examples of all methods in Figure~\ref{fig: examples}. In the first and second row, we can see that all base models tend to overestimate the restoration part, but this behavior is drastically reduced with test-time training. 
Row 3 shows an example with a very fine detail that is not predicted by any of the base models. While without test-time training the beak of the vase is ignored, our model can adapt to these details such that we obtain better results for the restoration shape. 
Cases like this happen quite often and do not change the reconstruction error significantly because the details are quite small, but qualitatively the preservation of such details makes a big difference.

%
A similar but more extreme effect can be observed in row 4 where the vase contains a plant, a case which is not covered in the training data. Due to the test-time training, we can adapt to this while the other methods cannot. 
The shape of the complete vase is not quite correct but there is also no information in the original fractured shape about the geometry of the lower part.

\subsection{Ablation Study}\label{subsec:ablation}

We evaluate the best latent dimensionality as well as the effect of the test-time training. 
The difference between using test-time training and not using it is reported in the complete results of \Cref{table: 10per} and \Cref{table: 50per}.

To evaluate the effect of the size of the latent space, we choose $\dim z_C=\dim z_B$ and train all models on the mugs object class. 
We find that the model performance decreases when we increase the dimensionality of the latent code drastically. 
One explanation for this observation could be that too many degrees of freedom prevent learning of class information and instead overfit on the training data. 
The results are reported in \Cref{tab:ablation}.
We used the best result of the ablation study, $\dim z_C=\dim z_B = 200$ for all our experiments.

\begin{table}[h!]
\centering 
\begin{tabular}{r|r|r||r}
$\dim z_C$&$\dim z_B$&$\dim z_C+z_B$ & CD \\ \hline
$128$&$64$&192 & 1.9 \\
$100$&$100$&$200$&  1.4\\
$200$&$200$&$400$&  \textbf{1.2}\\
$300$&$300$& $600$& 3.6\\
$400$&$400$&$800$&  3.1\\
$500$&$500$& $1000$& 3.8\\
$600$&$600$& $1200$& 3.9\\
$1000$&$1000$& $2000$&4.4\\
$1300$&$1300$&$2600$& 6.0 
\end{tabular}
\caption{Ablation study for the dimension of the latent Code.}
\label{tab:ablation}
\end{table}

%% file: chapters/07_conclusion.tex
\section{Conclusion} \label{sec:Conclusion}

We proposed a new framework to solve the 3D volumetric shape restoration problem by using test-time training. 
The problem of shape restoration requires a combination of broad class knowledge but also the ability to very accurately fit to the input shape which is a setting particularly suited for test-time training. 
A simple class-wise trained network does often dismiss small geometric details, but we are able to retain them through finetuning during inference. 
Additionally, we have proposed several network improvements for DeepMend~\cite{deepmend2022} and shown the effectiveness of our new pipeline on ShapeNet with two different levels of partiality.

%% file: main.bbl
\begin{thebibliography}{}

\bibitem [\protect \citeauthoryear {%
Anguelov%
\ \protect \BOthers {.}}{%
Anguelov%
\ \protect \BOthers {.}}{%
{\protect \APACyear {2005}}%
}]{%
shape_completion_anguelov}
\APACinsertmetastar {%
shape_completion_anguelov}%
\begin{APACrefauthors}%
Anguelov, D.%
, Srinivasan, P.%
, Koller, D.%
, Thrun, S.%
, Rodgers, J.%
\BCBL {}\ \BBA {} Davis, J.%
\end{APACrefauthors}%
\unskip\
\newblock
\APACrefYearMonthDay{2005}{}{}.
\newblock
{\BBOQ}\APACrefatitle {SCAPE: shape completion and animation of people} {Scape:
  shape completion and animation of people}.{\BBCQ}
\newblock
\BIn{} \APACrefbtitle {ACM SIGGRAPH 2005 Papers} {Acm siggraph 2005 papers}\
  (\BPG~408–416).
\newblock
\APACaddressPublisher{New York, NY, USA}{Association for Computing Machinery}.
\newblock
\begin{APACrefURL} \url{https://doi.org/10.1145/1186822.1073207}
  \end{APACrefURL}
\newblock
\doi{10.1145/1186822.1073207}
\PrintBackRefs{\CurrentBib}

\bibitem [\protect \citeauthoryear {%
Chang%
\ \protect \BOthers {.}}{%
Chang%
\ \protect \BOthers {.}}{%
{\protect \APACyear {2015}}%
}]{%
shapenet2015}
\APACinsertmetastar {%
shapenet2015}%
\begin{APACrefauthors}%
Chang, A\BPBI X.%
, Funkhouser, T.%
, Guibas, L.%
, Hanrahan, P.%
, Huang, Q.%
, Li, Z.%
\BDBL {}Yu, F.%
\end{APACrefauthors}%
\unskip\
\newblock
\APACrefYearMonthDay{2015}{}{}.
\newblock
\APACrefbtitle {{ShapeNet: An Information-Rich 3D Model Repository}}
  {{ShapeNet: An Information-Rich 3D Model Repository}}\
  \APACbVolEdTR{}{\BTR{}\ \BNUM\ arXiv:1512.03012 [cs.GR]}.
\PrintBackRefs{\CurrentBib}

\bibitem [\protect \citeauthoryear {%
Chibane%
\ \protect \BOthers {.}}{%
Chibane%
\ \protect \BOthers {.}}{%
{\protect \APACyear {2020}}%
}]{%
chibane20ifnet}
\APACinsertmetastar {%
chibane20ifnet}%
\begin{APACrefauthors}%
Chibane, J.%
, Alldieck, T.%
\BCBL {}\ \BBA {} Pons-Moll, G.%
\end{APACrefauthors}%
\unskip\
\newblock
\APACrefYearMonthDay{2020}{}{}.
\newblock
{\BBOQ}\APACrefatitle {Implicit Functions in Feature Space for 3D Shape
  Reconstruction and Completion} {Implicit functions in feature space for 3d
  shape reconstruction and completion}.{\BBCQ}
\newblock
\BIn{} \APACrefbtitle {{IEEE} Conference on Computer Vision and Pattern
  Recognition (CVPR).} {{IEEE} conference on computer vision and pattern
  recognition (cvpr).}
\PrintBackRefs{\CurrentBib}

\bibitem [\protect \citeauthoryear {%
Cui%
\ \protect \BOthers {.}}{%
Cui%
\ \protect \BOthers {.}}{%
{\protect \APACyear {2024}}%
}]{%
cui2024neusdfusion}
\APACinsertmetastar {%
cui2024neusdfusion}%
\begin{APACrefauthors}%
Cui, R.%
, Liu, W.%
, Sun, W.%
, Wang, S.%
, Shang, T.%
, Li, Y.%
\BDBL {}Ji, P.%
\end{APACrefauthors}%
\unskip\
\newblock
\APACrefYearMonthDay{2024}{}{}.
\newblock
{\BBOQ}\APACrefatitle {NeuSDFusion: A Spatial-Aware Generative Model for 3D
  Shape Completion, Reconstruction, and Generation} {Neusdfusion: A
  spatial-aware generative model for 3d shape completion, reconstruction, and
  generation}.{\BBCQ}
\newblock
\APACjournalVolNumPages{arXiv: 2403.18241}{}{}{}.
\PrintBackRefs{\CurrentBib}

\bibitem [\protect \citeauthoryear {%
Dai%
\ \protect \BOthers {.}}{%
Dai%
\ \protect \BOthers {.}}{%
{\protect \APACyear {2017}}%
}]{%
dai2017completion}
\APACinsertmetastar {%
dai2017completion}%
\begin{APACrefauthors}%
Dai, A.%
, Qi, C.%
\BCBL {}\ \BBA {} Niessner, M.%
\end{APACrefauthors}%
\unskip\
\newblock
\APACrefYearMonthDay{2017}{}{}.
\newblock
{\BBOQ}\APACrefatitle {Shape Completion using 3D-Encoder-Predictor CNNs and
  Shape Synthesis} {Shape completion using 3d-encoder-predictor cnns and shape
  synthesis}.{\BBCQ}
\newblock
\BIn{} \APACrefbtitle {{IEEE} Conference on Computer Vision and Pattern
  Recognition {(CVPR)}.} {{IEEE} conference on computer vision and pattern
  recognition {(CVPR)}.}
\PrintBackRefs{\CurrentBib}

\bibitem [\protect \citeauthoryear {%
Erkoç%
\ \protect \BOthers {.}}{%
Erkoç%
\ \protect \BOthers {.}}{%
{\protect \APACyear {2023}}%
}]{%
erkoc2023hyperdiff}
\APACinsertmetastar {%
erkoc2023hyperdiff}%
\begin{APACrefauthors}%
Erkoç, Z.%
, Ma, F.%
, Shan, Q.%
, Nießner, M.%
\BCBL {}\ \BBA {} Dai, A.%
\end{APACrefauthors}%
\unskip\
\newblock
\APACrefYearMonthDay{2023}{}{}.
\newblock
{\BBOQ}\APACrefatitle {HyperDiffusion: Generating Implicit Neural Fields with
  Weight-Space Diffusion} {Hyperdiffusion: Generating implicit neural fields
  with weight-space diffusion}.{\BBCQ}
\newblock
\BIn{} \APACrefbtitle {European Conference on Computer Vision {(ECCV)}.}
  {European conference on computer vision {(ECCV)}.}
\PrintBackRefs{\CurrentBib}

\bibitem [\protect \citeauthoryear {%
Gafencu%
\ \protect \BOthers {.}}{%
Gafencu%
\ \protect \BOthers {.}}{%
{\protect \APACyear {2024}}%
}]{%
gafencu2024shape}
\APACinsertmetastar {%
gafencu2024shape}%
\begin{APACrefauthors}%
Gafencu, M\BHBI A.%
, Velikova, Y.%
, Saleh, M.%
, Ungi, T.%
, Navab, N.%
, Wendler, T.%
\BCBL {}\ \BBA {} Azampour, M\BPBI F.%
\end{APACrefauthors}%
\unskip\
\newblock
\APACrefYearMonthDay{2024}{}{}.
\newblock
{\BBOQ}\APACrefatitle {Shape Completion in the Dark: Completing Vertebrae
  Morphology from 3D Ultrasound} {Shape completion in the dark: Completing
  vertebrae morphology from 3d ultrasound}.{\BBCQ}
\newblock
\APACjournalVolNumPages{arXiv: 2404.07668}{}{}{}.
\PrintBackRefs{\CurrentBib}

\bibitem [\protect \citeauthoryear {%
Galvis%
\ \protect \BOthers {.}}{%
Galvis%
\ \protect \BOthers {.}}{%
{\protect \APACyear {2024}}%
}]{%
galvis2024scdiff}
\APACinsertmetastar {%
galvis2024scdiff}%
\begin{APACrefauthors}%
Galvis, J\BPBI D.%
, Zuo, X.%
, Schaefer, S.%
\BCBL {}\ \BBA {} Leutengger, S.%
\end{APACrefauthors}%
\unskip\
\newblock
\APACrefYearMonthDay{2024}{}{}.
\newblock
{\BBOQ}\APACrefatitle {SC-Diff: 3D Shape Completion with Latent Diffusion
  Models} {Sc-diff: 3d shape completion with latent diffusion models}.{\BBCQ}
\newblock
\APACjournalVolNumPages{arXiv:2403.12470}{}{}{}.
\PrintBackRefs{\CurrentBib}

\bibitem [\protect \citeauthoryear {%
Gandelsman%
\ \protect \BOthers {.}}{%
Gandelsman%
\ \protect \BOthers {.}}{%
{\protect \APACyear {2022}}%
}]{%
gandelsman2022ttt}
\APACinsertmetastar {%
gandelsman2022ttt}%
\begin{APACrefauthors}%
Gandelsman, Y.%
, Sun, Y.%
, Chen, X.%
\BCBL {}\ \BBA {} Efros, A\BPBI A.%
\end{APACrefauthors}%
\unskip\
\newblock
\APACrefYearMonthDay{2022}{}{}.
\newblock
{\BBOQ}\APACrefatitle {Test-Time Training with Masked Autoencoders} {Test-time
  training with masked autoencoders}.{\BBCQ}
\newblock
\BIn{} \APACrefbtitle {Conference on Neural Information Processing Systems
  (NeurIPS).} {Conference on neural information processing systems (neurips).}
\PrintBackRefs{\CurrentBib}

\bibitem [\protect \citeauthoryear {%
Gao%
\ \protect \BOthers {.}}{%
Gao%
\ \protect \BOthers {.}}{%
{\protect \APACyear {2022}}%
}]{%
gao2022get3d}
\APACinsertmetastar {%
gao2022get3d}%
\begin{APACrefauthors}%
Gao, J.%
, Shen, T.%
, Wang, Z.%
, Chen, W.%
, Yin, K.%
, Li, D.%
\BDBL {}Fidler, S.%
\end{APACrefauthors}%
\unskip\
\newblock
\APACrefYearMonthDay{2022}{}{}.
\newblock
{\BBOQ}\APACrefatitle {GET3D: A Generative Model of High Quality 3D Textured
  Shapes Learned from Images} {Get3d: A generative model of high quality 3d
  textured shapes learned from images}.{\BBCQ}
\newblock
\BIn{} \APACrefbtitle {Advances In Neural Information Processing Systems.}
  {Advances in neural information processing systems.}
\PrintBackRefs{\CurrentBib}

\bibitem [\protect \citeauthoryear {%
Hanocka%
\ \protect \BOthers {.}}{%
Hanocka%
\ \protect \BOthers {.}}{%
{\protect \APACyear {2020}}%
}]{%
hanocka2020point2mesh}
\APACinsertmetastar {%
hanocka2020point2mesh}%
\begin{APACrefauthors}%
Hanocka, R.%
, Metzer, G.%
, Giryes, R.%
\BCBL {}\ \BBA {} Cohen-Or, D.%
\end{APACrefauthors}%
\unskip\
\newblock
\APACrefYearMonthDay{2020}{}{}.
\newblock
{\BBOQ}\APACrefatitle {Point2Mesh: A Self-Prior for Deformable Meshes}
  {Point2mesh: A self-prior for deformable meshes}.{\BBCQ}
\newblock
\BIn{} \APACrefbtitle {ACM Transactions on Graphics (Proc. SIGGRAPH).} {Acm
  transactions on graphics (proc. siggraph).}
\PrintBackRefs{\CurrentBib}

\bibitem [\protect \citeauthoryear {%
Hu%
\ \protect \BOthers {.}}{%
Hu%
\ \protect \BOthers {.}}{%
{\protect \APACyear {2024}}%
}]{%
hu2024topology}
\APACinsertmetastar {%
hu2024topology}%
\begin{APACrefauthors}%
Hu, J.%
, Fei, B.%
, Xu, B.%
, Hou, F.%
, Yang, W.%
, Wang, S.%
\BDBL {}He, Y.%
\end{APACrefauthors}%
\unskip\
\newblock
\APACrefYearMonthDay{2024}{}{}.
\newblock
{\BBOQ}\APACrefatitle {Topology-Aware Latent Diffusion for 3D Shape Generation}
  {Topology-aware latent diffusion for 3d shape generation}.{\BBCQ}
\newblock
\APACjournalVolNumPages{arXiv:2401.17603}{}{}{}.
\PrintBackRefs{\CurrentBib}

\bibitem [\protect \citeauthoryear {%
Kazhdan%
\ \protect \BOthers {.}}{%
Kazhdan%
\ \protect \BOthers {.}}{%
{\protect \APACyear {2006}}%
}]{%
Kazhdan:2006:poisson}
\APACinsertmetastar {%
Kazhdan:2006:poisson}%
\begin{APACrefauthors}%
Kazhdan, M\BPBI M.%
, Bolitho, M.%
\BCBL {}\ \BBA {} Hoppe, H.%
\end{APACrefauthors}%
\unskip\
\newblock
\APACrefYearMonthDay{2006}{}{}.
\newblock
{\BBOQ}\APACrefatitle {Poisson Surface Reconstruction} {Poisson surface
  reconstruction}.{\BBCQ}
\newblock
\BIn{} \APACrefbtitle {Proceedings of the Fourth Eurographics Symposium on
  Geometry Processing} {Proceedings of the fourth eurographics symposium on
  geometry processing}\ (\BVOL~256, \BPG~61-70).
\newblock
\APACaddressPublisher{}{Eurographics Association}.
\PrintBackRefs{\CurrentBib}

\bibitem [\protect \citeauthoryear {%
Lamb%
\ \protect \BOthers {.}}{%
Lamb%
\ \protect \BOthers {.}}{%
{\protect \APACyear {2022}}%
{\protect \APACexlab {{\protect \BCnt {1}}}}}]{%
lamb2022deepjoin}
\APACinsertmetastar {%
lamb2022deepjoin}%
\begin{APACrefauthors}%
Lamb, N.%
, Banerjee, S.%
\BCBL {}\ \BBA {} Banerjee, N\BPBI K.%
\end{APACrefauthors}%
\unskip\
\newblock
\APACrefYearMonthDay{2022{\protect \BCnt {1}}}{}{}.
\newblock
{\BBOQ}\APACrefatitle {DeepJoin: Learning a Joint Occupancy, Signed Distance,
  and Normal Field Function for Shape Repair} {Deepjoin: Learning a joint
  occupancy, signed distance, and normal field function for shape
  repair}.{\BBCQ}
\newblock
\APACjournalVolNumPages{ACM Transactions on Graphics (TOG)}{41}{6}{1--10}.
\PrintBackRefs{\CurrentBib}

\bibitem [\protect \citeauthoryear {%
Lamb%
\ \protect \BOthers {.}}{%
Lamb%
\ \protect \BOthers {.}}{%
{\protect \APACyear {2022}}%
{\protect \APACexlab {{\protect \BCnt {2}}}}}]{%
deepmend2022}
\APACinsertmetastar {%
deepmend2022}%
\begin{APACrefauthors}%
Lamb, N.%
, Banerjee, S.%
\BCBL {}\ \BBA {} Banerjee, N\BPBI K.%
\end{APACrefauthors}%
\unskip\
\newblock
\APACrefYearMonthDay{2022{\protect \BCnt {2}}}{}{}.
\newblock
{\BBOQ}\APACrefatitle {Deep{M}end: Learning Occupancy Functions to Represent
  Shape for Repair} {Deep{M}end: Learning occupancy functions to represent
  shape for repair}.{\BBCQ}
\newblock
\BIn{} \APACrefbtitle {European Conference on Computer Vision {(ECCV)}.}
  {European conference on computer vision {(ECCV)}.}
\PrintBackRefs{\CurrentBib}

\bibitem [\protect \citeauthoryear {%
Lamb%
\ \protect \BOthers {.}}{%
Lamb%
\ \protect \BOthers {.}}{%
{\protect \APACyear {2023}}%
}]{%
fantasticbreaks}
\APACinsertmetastar {%
fantasticbreaks}%
\begin{APACrefauthors}%
Lamb, N.%
, Palmer, C.%
, Molloy, B.%
, Banerjee, S.%
\BCBL {}\ \BBA {} Kholgade~Banerjee, N.%
\end{APACrefauthors}%
\unskip\
\newblock
\APACrefYearMonthDay{2023}{June}{}.
\newblock
{\BBOQ}\APACrefatitle {Fantastic Breaks: A Dataset of Paired 3D Scans of
  Real-World Broken Objects and Their Complete Counterparts} {Fantastic breaks:
  A dataset of paired 3d scans of real-world broken objects and their complete
  counterparts}.{\BBCQ}
\newblock
\BIn{} \APACrefbtitle {Proceedings of the IEEE/CVF Conference on Computer
  Vision and Pattern Recognition (CVPR).} {Proceedings of the ieee/cvf
  conference on computer vision and pattern recognition (cvpr).}
\PrintBackRefs{\CurrentBib}

\bibitem [\protect \citeauthoryear {%
Lamb%
\ \protect \BOthers {.}}{%
Lamb%
\ \protect \BOthers {.}}{%
{\protect \APACyear {2021}}%
}]{%
CreateFracture2021}
\APACinsertmetastar {%
CreateFracture2021}%
\begin{APACrefauthors}%
Lamb, N.%
, Wiederhold, N.%
, Lamb, B.%
, Banerjee, S.%
\BCBL {}\ \BBA {} Banerjee, N\BPBI K.%
\end{APACrefauthors}%
\unskip\
\newblock
\APACrefYearMonthDay{2021}{}{}.
\newblock
{\BBOQ}\APACrefatitle {Using Learned Visual and Geometric Features to Retrieve
  Complete 3D Proxies for Broken Objects} {Using learned visual and geometric
  features to retrieve complete 3d proxies for broken objects}.{\BBCQ}
\newblock
\BIn{} \APACrefbtitle {Symposium on Computational Fabrication} {Symposium on
  computational fabrication}\ (\BPGS\ 1--15).
\PrintBackRefs{\CurrentBib}

\bibitem [\protect \citeauthoryear {%
Lin%
\ \protect \BOthers {.}}{%
Lin%
\ \protect \BOthers {.}}{%
{\protect \APACyear {2023}}%
}]{%
lin2023magic3d}
\APACinsertmetastar {%
lin2023magic3d}%
\begin{APACrefauthors}%
Lin, C\BHBI H.%
, Gao, J.%
, Tang, L.%
, Takikawa, T.%
, Zeng, X.%
, Huang, X.%
\BDBL {}Lin, T\BHBI Y.%
\end{APACrefauthors}%
\unskip\
\newblock
\APACrefYearMonthDay{2023}{}{}.
\newblock
{\BBOQ}\APACrefatitle {Magic3D: High-Resolution Text-to-3D Content Creation}
  {Magic3d: High-resolution text-to-3d content creation}.{\BBCQ}
\newblock
\BIn{} \APACrefbtitle {IEEE Conference on Computer Vision and Pattern
  Recognition ({CVPR}).} {Ieee conference on computer vision and pattern
  recognition ({CVPR}).}
\PrintBackRefs{\CurrentBib}

\bibitem [\protect \citeauthoryear {%
Liu%
\ \protect \BOthers {.}}{%
Liu%
\ \protect \BOthers {.}}{%
{\protect \APACyear {2023}}%
}]{%
liu2023one2345}
\APACinsertmetastar {%
liu2023one2345}%
\begin{APACrefauthors}%
Liu, M.%
, Xu, C.%
, Jin, H.%
, Chen, L.%
, Xu, Z.%
, Su, H.%
\BCBL {}\ \BOthersPeriod {.}\end{APACrefauthors}%
\unskip\
\newblock
\APACrefYearMonthDay{2023}{}{}.
\newblock
{\BBOQ}\APACrefatitle {One-2-3-45: Any single image to 3d mesh in 45 seconds
  without per-shape optimization} {One-2-3-45: Any single image to 3d mesh in
  45 seconds without per-shape optimization}.{\BBCQ}
\newblock
\APACjournalVolNumPages{Advances In Neural Information Processing
  Systems}{}{}{}.
\PrintBackRefs{\CurrentBib}

\bibitem [\protect \citeauthoryear {%
Loiseau%
\ \protect \BOthers {.}}{%
Loiseau%
\ \protect \BOthers {.}}{%
{\protect \APACyear {2021}}%
}]{%
loiseau2021linearmodels}
\APACinsertmetastar {%
loiseau2021linearmodels}%
\begin{APACrefauthors}%
Loiseau, R.%
, Monnier, T.%
, Aubry, M.%
\BCBL {}\ \BBA {} Landrieu, L.%
\end{APACrefauthors}%
\unskip\
\newblock
\APACrefYearMonthDay{2021}{}{}.
\newblock
{\BBOQ}\APACrefatitle {Representing Shape Collections With Alignment-Aware
  Linear Models} {Representing shape collections with alignment-aware linear
  models}.{\BBCQ}
\newblock
\BIn{} \APACrefbtitle {International Conference on 3D Vision (3DV).}
  {International conference on 3d vision (3dv).}
\PrintBackRefs{\CurrentBib}

\bibitem [\protect \citeauthoryear {%
Lorensen%
\ \BBA {} Cline%
}{%
Lorensen%
\ \BBA {} Cline%
}{%
{\protect \APACyear {1987}}%
}]{%
MarchingCubes}
\APACinsertmetastar {%
MarchingCubes}%
\begin{APACrefauthors}%
Lorensen, W.%
\BCBT {}\ \BBA {} Cline, H.%
\end{APACrefauthors}%
\unskip\
\newblock
\APACrefYearMonthDay{1987}{08}{}.
\newblock
{\BBOQ}\APACrefatitle {Marching Cubes: A High Resolution 3D Surface
  Construction Algorithm} {Marching cubes: A high resolution 3d surface
  construction algorithm}.{\BBCQ}
\newblock
\APACjournalVolNumPages{ACM SIGGRAPH Computer Graphics}{21}{}{163-}.
\newblock
\doi{10.1145/37401.37422}
\PrintBackRefs{\CurrentBib}

\bibitem [\protect \citeauthoryear {%
Luo%
\ \BBA {} Hu%
}{%
Luo%
\ \BBA {} Hu%
}{%
{\protect \APACyear {2021}}%
}]{%
luo2021probpcs}
\APACinsertmetastar {%
luo2021probpcs}%
\begin{APACrefauthors}%
Luo, S.%
\BCBT {}\ \BBA {} Hu, W.%
\end{APACrefauthors}%
\unskip\
\newblock
\APACrefYearMonthDay{2021}{}{}.
\newblock
{\BBOQ}\APACrefatitle {Diffusion probabilistic models for 3d point cloud
  generation} {Diffusion probabilistic models for 3d point cloud
  generation}.{\BBCQ}
\newblock
\BIn{} \APACrefbtitle {IEEE/CVF Conference on Computer Vision and Pattern
  Recognition (CVPR).} {Ieee/cvf conference on computer vision and pattern
  recognition (cvpr).}
\PrintBackRefs{\CurrentBib}

\bibitem [\protect \citeauthoryear {%
Mehta%
\ \protect \BOthers {.}}{%
Mehta%
\ \protect \BOthers {.}}{%
{\protect \APACyear {2022}}%
}]{%
Mehta2022levelset}
\APACinsertmetastar {%
Mehta2022levelset}%
\begin{APACrefauthors}%
Mehta, I.%
, Chandraker, M.%
\BCBL {}\ \BBA {} Ramamoorthi, R.%
\end{APACrefauthors}%
\unskip\
\newblock
\APACrefYearMonthDay{2022}{}{}.
\newblock
{\BBOQ}\APACrefatitle {A Level Set Theory for Neural Implicit Evolution Under
  Explicit Flows} {A level set theory for neural implicit evolution under
  explicit flows}.{\BBCQ}
\newblock
\BIn{} \APACrefbtitle {Computer Vision – ECCV 2022: 17th European Conference,
  Tel Aviv, Israel, October 23–27, 2022, Proceedings, Part II} {Computer
  vision – eccv 2022: 17th european conference, tel aviv, israel, october
  23–27, 2022, proceedings, part ii}\ (\BPG~711–729).
\PrintBackRefs{\CurrentBib}

\bibitem [\protect \citeauthoryear {%
Park%
\ \protect \BOthers {.}}{%
Park%
\ \protect \BOthers {.}}{%
{\protect \APACyear {2019}}%
}]{%
DeepSDF}
\APACinsertmetastar {%
DeepSDF}%
\begin{APACrefauthors}%
Park, J\BPBI J.%
, Florence, P.%
, Straub, J.%
, Newcombe, R.%
\BCBL {}\ \BBA {} Lovegrove, S.%
\end{APACrefauthors}%
\unskip\
\newblock
\APACrefYearMonthDay{2019}{June}{}.
\newblock
{\BBOQ}\APACrefatitle {Deep{SDF}: Learning Continuous Signed Distance Functions
  for Shape Representation} {Deep{SDF}: Learning continuous signed distance
  functions for shape representation}.{\BBCQ}
\newblock
\BIn{} \APACrefbtitle {Proceedings of the IEEE/CVF Conference on Computer
  Vision and Pattern Recognition (CVPR).} {Proceedings of the ieee/cvf
  conference on computer vision and pattern recognition (cvpr).}
\PrintBackRefs{\CurrentBib}

\bibitem [\protect \citeauthoryear {%
Poursaeed%
\ \protect \BOthers {.}}{%
Poursaeed%
\ \protect \BOthers {.}}{%
{\protect \APACyear {2020}}%
}]{%
poursaeed2020coupling}
\APACinsertmetastar {%
poursaeed2020coupling}%
\begin{APACrefauthors}%
Poursaeed, O.%
, Fisher, M.%
, Aigerman, N.%
\BCBL {}\ \BBA {} Kim, V\BPBI G.%
\end{APACrefauthors}%
\unskip\
\newblock
\APACrefYearMonthDay{2020}{}{}.
\newblock
{\BBOQ}\APACrefatitle {Coupling explicit and implicit surface representations
  for generative 3d modeling} {Coupling explicit and implicit surface
  representations for generative 3d modeling}.{\BBCQ}
\newblock
\BIn{} \APACrefbtitle {European Conference on Computer Vision} {European
  conference on computer vision}\ (\BPGS\ 667--683).
\PrintBackRefs{\CurrentBib}

\bibitem [\protect \citeauthoryear {%
Qiu%
\ \protect \BOthers {.}}{%
Qiu%
\ \protect \BOthers {.}}{%
{\protect \APACyear {2023}}%
}]{%
qiu2023richdreamer}
\APACinsertmetastar {%
qiu2023richdreamer}%
\begin{APACrefauthors}%
Qiu, L.%
, Chen, G.%
, Gu, X.%
, zuo, Q.%
, Xu, M.%
, Wu, Y.%
\BDBL {}Han, X.%
\end{APACrefauthors}%
\unskip\
\newblock
\APACrefYearMonthDay{2023}{}{}.
\newblock
{\BBOQ}\APACrefatitle {RichDreamer: A Generalizable Normal-Depth Diffusion
  Model for Detail Richness in Text-to-3D} {Richdreamer: A generalizable
  normal-depth diffusion model for detail richness in text-to-3d}.{\BBCQ}
\newblock
\APACjournalVolNumPages{IEEE/CVF Conference on Computer Vision and Pattern
  Recognition (CVPR)}{}{}{}.
\PrintBackRefs{\CurrentBib}

\bibitem [\protect \citeauthoryear {%
Sain%
\ \protect \BOthers {.}}{%
Sain%
\ \protect \BOthers {.}}{%
{\protect \APACyear {2022}}%
}]{%
sain2022sketch}
\APACinsertmetastar {%
sain2022sketch}%
\begin{APACrefauthors}%
Sain, A.%
, Kumar, A.%
, Potlapalli, B\BPBI V.%
, Chowdhury, P\BPBI N.%
, Xiang, T.%
\BCBL {}\ \BBA {} Song, Y\BHBI Z.%
\end{APACrefauthors}%
\unskip\
\newblock
\APACrefYearMonthDay{2022}{}{}.
\newblock
{\BBOQ}\APACrefatitle {Sketch3T: Test-Time Training for Zero-Shot SBIR}
  {Sketch3t: Test-time training for zero-shot sbir}.{\BBCQ}
\newblock
\BIn{} \APACrefbtitle {IEEE/CVF Conference on Computer Vision and Pattern
  Recognition (CVPR).} {Ieee/cvf conference on computer vision and pattern
  recognition (cvpr).}
\PrintBackRefs{\CurrentBib}

\bibitem [\protect \citeauthoryear {%
Sellán%
\ \protect \BOthers {.}}{%
Sellán%
\ \protect \BOthers {.}}{%
{\protect \APACyear {2022}}%
}]{%
breakingbad}
\APACinsertmetastar {%
breakingbad}%
\begin{APACrefauthors}%
Sellán, S.%
, Chen, Y\BHBI C.%
, Wu, Z.%
, Garg, A.%
\BCBL {}\ \BBA {} Jacobson, A.%
\end{APACrefauthors}%
\unskip\
\newblock
\APACrefYearMonthDay{2022}{}{}.
\newblock
{\BBOQ}\APACrefatitle {Breaking Bad: A Dataset for Geometric Fracture and
  Reassembly} {Breaking bad: A dataset for geometric fracture and
  reassembly}.{\BBCQ}
\newblock
\BIn{} \APACrefbtitle {Conference on Neural Information Processing Systems
  (NeurIPS).} {Conference on neural information processing systems (neurips).}
\PrintBackRefs{\CurrentBib}

\bibitem [\protect \citeauthoryear {%
Sharf%
\ \protect \BOthers {.}}{%
Sharf%
\ \protect \BOthers {.}}{%
{\protect \APACyear {2004}}%
}]{%
context-based_surface}
\APACinsertmetastar {%
context-based_surface}%
\begin{APACrefauthors}%
Sharf, A.%
, Alexa, M.%
\BCBL {}\ \BBA {} Cohen-Or, D.%
\end{APACrefauthors}%
\unskip\
\newblock
\APACrefYearMonthDay{2004}{}{}.
\newblock
{\BBOQ}\APACrefatitle {Context-based surface completion} {Context-based surface
  completion}.{\BBCQ}
\newblock
\BIn{} \APACrefbtitle {ACM SIGGRAPH 2004 Papers} {Acm siggraph 2004 papers}\
  (\BPGS\ 878--887).
\PrintBackRefs{\CurrentBib}

\bibitem [\protect \citeauthoryear {%
Slavcheva%
\ \protect \BOthers {.}}{%
Slavcheva%
\ \protect \BOthers {.}}{%
{\protect \APACyear {2017}}%
}]{%
slavcheva2017killing}
\APACinsertmetastar {%
slavcheva2017killing}%
\begin{APACrefauthors}%
Slavcheva, M.%
, Baust, M.%
, Cremers, D.%
\BCBL {}\ \BBA {} Ilic, S.%
\end{APACrefauthors}%
\unskip\
\newblock
\APACrefYearMonthDay{2017}{}{}.
\newblock
{\BBOQ}\APACrefatitle {KillingFusion: Non-rigid 3D Reconstruction without
  Correspondences} {Killingfusion: Non-rigid 3d reconstruction without
  correspondences}.{\BBCQ}
\newblock
\BIn{} \APACrefbtitle {2017 IEEE Conference on Computer Vision and Pattern
  Recognition (CVPR).} {2017 ieee conference on computer vision and pattern
  recognition (cvpr).}
\PrintBackRefs{\CurrentBib}

\bibitem [\protect \citeauthoryear {%
B.~Sun%
\ \protect \BOthers {.}}{%
B.~Sun%
\ \protect \BOthers {.}}{%
{\protect \APACyear {2022}}%
}]{%
bo2022patchrd}
\APACinsertmetastar {%
bo2022patchrd}%
\begin{APACrefauthors}%
Sun, B.%
, Kim, V.%
, Aigerman, N.%
, Qixing, H.%
\BCBL {}\ \BBA {} Chaudhuri, S.%
\end{APACrefauthors}%
\unskip\
\newblock
\APACrefYearMonthDay{2022}{}{}.
\newblock
{\BBOQ}\APACrefatitle {PatchRD: Detail-Preserving Shape Completion by Learning
  Patch Retrieval and Deformation} {Patchrd: Detail-preserving shape completion
  by learning patch retrieval and deformation}.{\BBCQ}
\newblock
\BIn{} \APACrefbtitle {European Conference on Computer Vision {(ECCV)}.}
  {European conference on computer vision {(ECCV)}.}
\PrintBackRefs{\CurrentBib}

\bibitem [\protect \citeauthoryear {%
Y.~Sun%
\ \protect \BOthers {.}}{%
Y.~Sun%
\ \protect \BOthers {.}}{%
{\protect \APACyear {2020}}%
}]{%
sun2020ttt}
\APACinsertmetastar {%
sun2020ttt}%
\begin{APACrefauthors}%
Sun, Y.%
, Wang, X.%
, Liu, Z.%
, Miller, J.%
, Efros, A\BPBI A.%
\BCBL {}\ \BBA {} Hardt, M.%
\end{APACrefauthors}%
\unskip\
\newblock
\APACrefYearMonthDay{2020}{}{}.
\newblock
{\BBOQ}\APACrefatitle {Test-Time Training with Self-Supervision for
  Generalization under Distribution Shifts} {Test-time training with
  self-supervision for generalization under distribution shifts}.{\BBCQ}
\newblock
\BIn{} \APACrefbtitle {International Conference on Machine Learning (ICML).}
  {International conference on machine learning (icml).}
\PrintBackRefs{\CurrentBib}

\bibitem [\protect \citeauthoryear {%
Wu%
\ \protect \BOthers {.}}{%
Wu%
\ \protect \BOthers {.}}{%
{\protect \APACyear {2015}}%
}]{%
wu20153d}
\APACinsertmetastar {%
wu20153d}%
\begin{APACrefauthors}%
Wu, Z.%
, Song, S.%
, Khosla, A.%
, Yu, F.%
, Zhang, L.%
, Tang, X.%
\BCBL {}\ \BBA {} Xiao, J.%
\end{APACrefauthors}%
\unskip\
\newblock
\APACrefYearMonthDay{2015}{}{}.
\newblock
{\BBOQ}\APACrefatitle {3d shapenets: A deep representation for volumetric
  shapes} {3d shapenets: A deep representation for volumetric shapes}.{\BBCQ}
\newblock
\BIn{} \APACrefbtitle {Proceedings of the IEEE conference on computer vision
  and pattern recognition} {Proceedings of the ieee conference on computer
  vision and pattern recognition}\ (\BPGS\ 1912--1920).
\PrintBackRefs{\CurrentBib}

\bibitem [\protect \citeauthoryear {%
Zeng%
\ \protect \BOthers {.}}{%
Zeng%
\ \protect \BOthers {.}}{%
{\protect \APACyear {2022}}%
}]{%
zeng2022lion}
\APACinsertmetastar {%
zeng2022lion}%
\begin{APACrefauthors}%
Zeng, X.%
, Vahdat, A.%
, Williams, F.%
, Gojcic, Z.%
, Litany, O.%
, Fidler, S.%
\BCBL {}\ \BBA {} Kreis, K.%
\end{APACrefauthors}%
\unskip\
\newblock
\APACrefYearMonthDay{2022}{}{}.
\newblock
{\BBOQ}\APACrefatitle {LION: Latent Point Diffusion Models for 3D Shape
  Generation} {Lion: Latent point diffusion models for 3d shape
  generation}.{\BBCQ}
\newblock
\BIn{} \APACrefbtitle {Advances in Neural Information Processing Systems
  (NeurIPS).} {Advances in neural information processing systems (neurips).}
\PrintBackRefs{\CurrentBib}

\bibitem [\protect \citeauthoryear {%
Zhu%
\ \protect \BOthers {.}}{%
Zhu%
\ \protect \BOthers {.}}{%
{\protect \APACyear {2023}}%
}]{%
zhu2023svdformer}
\APACinsertmetastar {%
zhu2023svdformer}%
\begin{APACrefauthors}%
Zhu, Z.%
, Chen, H.%
, He, X.%
, Wang, W.%
, Qin, J.%
\BCBL {}\ \BBA {} Wei, M.%
\end{APACrefauthors}%
\unskip\
\newblock
\APACrefYearMonthDay{2023}{}{}.
\newblock
{\BBOQ}\APACrefatitle {SVDFormer: Complementing Point Cloud via Self-view
  Augmentation and Self-structure Dual-generator} {Svdformer: Complementing
  point cloud via self-view augmentation and self-structure
  dual-generator}.{\BBCQ}
\newblock
\APACjournalVolNumPages{arXiv:2307.08492}{}{}{}.
\PrintBackRefs{\CurrentBib}

\bibitem [\protect \citeauthoryear {%
Zuffi%
\ \protect \BOthers {.}}{%
Zuffi%
\ \protect \BOthers {.}}{%
{\protect \APACyear {2018}}%
}]{%
zuffi2018smal}
\APACinsertmetastar {%
zuffi2018smal}%
\begin{APACrefauthors}%
Zuffi, S.%
, Kanazawa, A.%
\BCBL {}\ \BBA {} Black, M\BPBI J.%
\end{APACrefauthors}%
\unskip\
\newblock
\APACrefYearMonthDay{2018}{}{}.
\newblock
{\BBOQ}\APACrefatitle {Lions and Tigers and Bears: Capturing Non-Rigid, 3D,
  Articulated Shape from Images} {Lions and tigers and bears: Capturing
  non-rigid, 3d, articulated shape from images}.{\BBCQ}
\newblock
\BIn{} \APACrefbtitle {IEEE/CVF Conference on Computer Vision and Pattern
  Recognition (CVPR).} {Ieee/cvf conference on computer vision and pattern
  recognition (cvpr).}
\PrintBackRefs{\CurrentBib}

\end{thebibliography}
